\documentclass{article}

\usepackage[final]{neurips_2025}

\usepackage[utf8]{inputenc} % allow utf-8 input
\usepackage[T1]{fontenc}    % use 8-bit T1 fonts
\usepackage{hyperref}       % hyperlinks
\usepackage{url}            % simple URL typesetting
\usepackage{booktabs}       % professional-quality tables
\usepackage{amsfonts}       % blackboard math symbols
\usepackage{nicefrac}       % compact symbols for 1/2, etc.
\usepackage{microtype}      % microtypography
\usepackage{xcolor}         % colors
\usepackage[most]{tcolorbox}
\tcbuselibrary{breakable}
\tcbuselibrary{skins}

\usepackage{graphicx}
\usepackage{multirow}
\usepackage{amsmath}
\usepackage{colortbl}
\usepackage{wrapfig}
\usepackage{pifont}% http://ctan.org/pkg/pifont
\newcommand{\cmark}{\ding{51}}%
\newcommand{\xmark}{\ding{55}}%
\usepackage{float}
\usepackage{adjustbox}
\usepackage{tabularx}
\usepackage{wrapfig}
\usepackage{caption}
\usepackage{enumitem}
\usepackage{fix-cm}

\title{WritingBench: A Comprehensive Benchmark for Generative Writing}

\author{
    \textbf{Yuning Wu\textsuperscript{1}},
    \textbf{Jiahao Mei\textsuperscript{1,3}},
    \textbf{Ming Yan\textsuperscript{1}\thanks{Corresponding authors}},
    \textbf{Chenliang Li\textsuperscript{1}},
    \textbf{Shaopeng Lai\textsuperscript{1}},
    \textbf{Yuran Ren\textsuperscript{2}}, 
    \\
    \textbf{Zijia Wang\textsuperscript{2}},
    \textbf{Ji Zhang\textsuperscript{1}},
    \textbf{Mengyue Wu\textsuperscript{3}},
    \textbf{Qin Jin}\textsuperscript{\textbf{2}*},
    \textbf{Fei Huang\textsuperscript{1}}
    \\
    \textsuperscript{1}Alibaba Group,
    \textsuperscript{2}Renmin University of China,
    \textsuperscript{3}Shanghai Jiao Tong University
    \\
    \small{
        \{yuningwu, ym119608\}@alibaba-inc.com, qjin@ruc.edu.cn
    }
}

\begin{document}

\maketitle

\begin{abstract}
Recent advancements in large language models (LLMs) have significantly enhanced text generation capabilities, yet evaluating their performance in generative writing remains a challenge. 
Existing benchmarks primarily focus on generic text generation or limited in writing tasks, failing to capture the diverse requirements of high-quality written contents across various domains. 
To bridge this gap, we present \textbf{WritingBench}, a comprehensive benchmark designed to evaluate LLMs across 6 core writing domains and 100 subdomains.%, encompassing creative, persuasive, informative, and technical writing.
We further propose a \textit{query-dependent evaluation} framework that empowers LLMs to dynamically generate instance-specific assessment criteria. This framework is complemented by a fine-tuned critic model for criteria-aware scoring, enabling evaluations in style, format and length. The framework's validity is further demonstrated by its data curation capability, which enables a 7B-parameter model to outperform the performance of GPT-4o in writing. We open-source the benchmark, along with evaluation tools and modular framework components, to advance the development of LLMs in writing.

\vspace{3pt}
\textbf{GitHub Repo:} https://github.com/X-PLUG/WritingBench

\vspace{-2pt}
\textbf{Leaderboard:} https://huggingface.co/spaces/WritingBench/WritingBench

\end{abstract}

\section{Introduction}

% \begin{figure}[h]
%   \centering
%   \includegraphics[width=0.5\columnwidth]{pics/query.pdf}
%   \caption{WritingBench query example with color-coded requirements. The three black-bordered categories highlight essential requirements analyzed in follow-up assessments. Red phrases correlate with gray-shaded writing support materials.}
%   \label{fig:writing query}
% \end{figure}

% The rapid evolution of large language models (LLMs) has significantly advanced machine-generated writing capabilities~\citep{XXX,survey}. However, the growing diversity of real-world user demands poses substantial challenges for systematic evaluation. First, writing tasks now encompass an unprecedented breadth of domains, ranging from creative writing to educational content creation, as LLMs attract users across industries and demographics, tackling previously intractable writing scenarios. Second, user requirements have grown increasingly granular, demanding precise control over stylistic adaptation, structural coherence, and contextual fidelity. Furthermore, real-world writing tasks frequently involve lengthy source materials, resulting in complex queries that extend far beyond simple prompts. These factors-domain diversity, requirement complexity, and material processing-render conventional evaluation frameworks inadequate, as narrow-domain simple-instruction benchmarks and static criteria fail to capture authentic writing requirements.

\begin{wrapfigure}{r}{0.46\textwidth}
  \vspace{-0.58cm}
  \centering
  \includegraphics[width=0.48\textwidth]{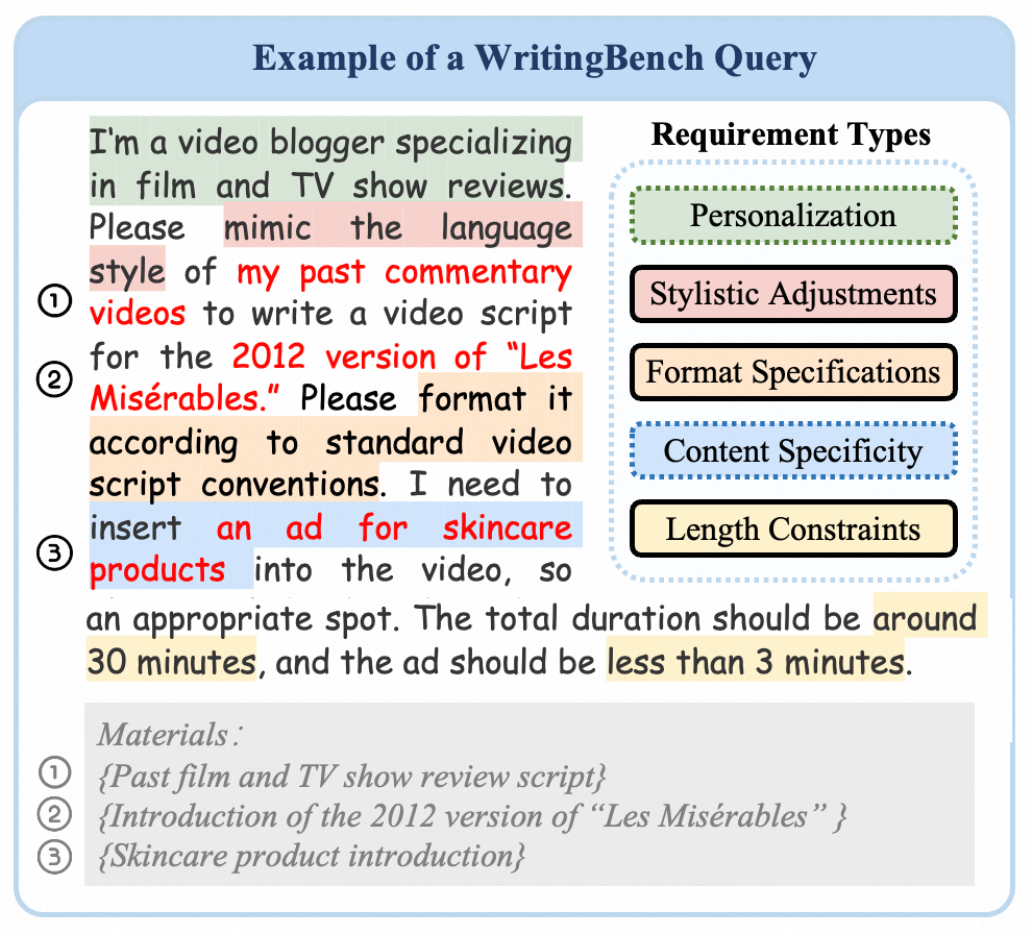}
  \vspace{-0.52cm}
  \caption{Query example with color-coded requirements. %Black-bordered categories highlight key assessment criteria;
  Red phrases correlate with gray-shaded writing support materials.}
  \vspace{-1.3cm}
  \label{fig:writing query}
\end{wrapfigure}

In recent years, LLMs~\citep{r1, 4ol} have revolutionized text generation, demonstrating impressive performance across diverse applications, from generating creative content~\citep{theatre-human, ex3} and assisting in education~\citep{education-survey1, education-survey2} to enhancing professional workflows~\citep{storm, metareview}. However, generative writing, which requires high levels of coherence, creativity, logical reasoning, and stylistic precision, poses a unique challenge that existing evaluation methods fail to address adequately. 
%As LLMs play an increasingly prominent role in these domains, establishing comprehensive and reliable benchmarks is crucial for evaluating their current performance and guiding future improvements in writing proficiency.

%Existing evaluation benchmarks exhibit two significant limitations. 

\begin{figure}[ht]
  \centering
  \includegraphics[width=1\textwidth]{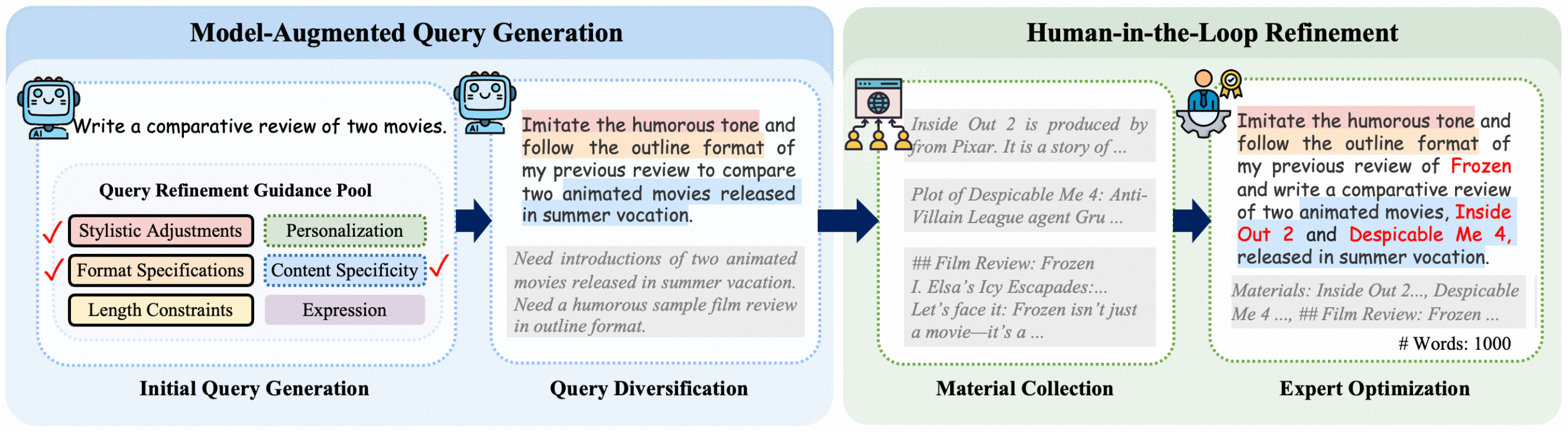}
  \caption{Construction pipeline of WritingBench queries. The refinement pool contains five writing requirements (three core competencies with black borders) and one expression type (purple). Checked strategies refine initial queries into multi-requirement prompts (color-coded text) with red phrases referencing gray materials. Implementation details in Section~\ref{ssec: benchmark_construction}.}
  \label{fig:Benchmark Construction}
  \vspace{-0.65em}
\end{figure}

Current evaluation benchmarks for generative writing suffer from two major limitations: 1) Limited scope and diversity in task formulation; and 2) Inadequate evaluation metrics for complex writing tasks. 
First, there is a significant lack of specialized benchmarks that cover a broad range of writing scenarios. Most existing writing-oriented benchmarks are restricted to single domains, such as fictions~\citep{nocha, confederacy-human}, and their task formulations tend to be simplistic---often relying on single-sentence prompts~\citep{longwriter} or a small set of instruction templates~\citep{eqbench, hellobench}. Additionally, many benchmarks use homogeneous input materials~\citep{hellobench, nocha}, limiting their ability to accommodate the complex and customized requirements inherent in real-world writing. As a result, they fail to capture the diversity and intricacies of practical writing tasks (see Figure~\ref{fig:writing query}).
Second, current automatic evaluation metrics lack the robustness needed for a comprehensive and nuanced assessment of writing quality. While LLM-based evaluation methods show promise in capturing semantic meanings~\citep{storm, hellobench, longwriter}, they typically rely on a narrow set of predefined criteria (e.g., fluency and coherence). As LLMs continue to evolve with increasingly sophisticated writing capabilities, these static evaluation criteria and frameworks are inadequate for assessing the complex, multi-dimensional nature of writing, including creativity, argumentation strength, and domain-specific adherence.

To address these challenges, we introduce \textbf{WritingBench}, a comprehensive benchmark and robust framework for evaluating general-purpose writing. Our approach begins with a carefully designed secondary domain categorization, grounded in real-world writing needs. We develop a four-stage query construction pipeline (illustrated in Figure~\ref{fig:Benchmark Construction}), where LLMs first generate and diversify writing queries, followed by human-driven material collection and optimization. This process ensures a diverse set of writing tasks with broad domain coverage, varied requirements, and integration of heterogeneous source of materials. %results in a set of writing queries that are characterized by broad domain coverage, varied requirements, and the integration of materials from diverse sources. 
%To facilitate a more nuanced evaluation of generated responses across different domains, 
To enable a more nuanced evaluation, we propose a \textit{query-dependent evaluation} framework where LLMs dynamically generates five instance-specific criteria, which are then scored by a fine-tuned critic model. 
Finally, we integrate the framework to filter writing-specific data and train a small-scale model to verify its ability in identifying high-quality writing samples.
% Crucially, the framework proves effective through data curation: filtering training data with our evaluation method enables 7B models to match SOTA performance, demonstrating its ability to identify high-quality writing samples.
Our primary contributions are as follows:
\begin{itemize}[left=0pt]
    \item We present \textbf{WritingBench}, an open-source writing benchmark comprising \textit{1,000} queries across \textit{6} primary domains and \textit{100} subdomains, featuring \textit{style}, \textit{format} and \textit{length} requirements. WritingBench supports extended-context generation with input ranging from tens to thousands of words, addressing real-world diversity. It facilitates systematic evaluation to identify improvement areas and highlights the potential of chain-of-thought (CoT) processes in creative tasks.
    \item We propose a \textit{query-dependent evaluation} framework that integrates instance-specific criteria generation with a criteria-aware scoring model. It achieves 84\% human alignment, significantly surpassing static-criteria baselines (67\%, 58\%). The effectiveness is further evidenced by its data curation capability-models trained with framework-filtered data outperform GPT-4o in writing.
    \item We publicly release WritingBench, including its evaluation protocols, criteria generation tools with an integrated critic model, and writing-enhanced models, to foster further research.  Available at: \url{https://github.com/X-PLUG/WritingBench}.
\end{itemize}

% $\bullet$ We present \textbf{WritingBench}, an open-source writing benchmark comprising \textit{1,000} queries across \textit{6} primary domains and \textit{100} subdomains,featuring \textit{style}, \textit{format} and \textit{length} requirements. WritingBench supports extended-context generation with input ranging from tens to thousands of words, addressing real-world diversity. It facilitates systematic evaluation to identify improvement areas and highlights the potential of chain-of-thought (CoT) processes in creative tasks.

% $\bullet$ We propose a \textit{query-dependent evaluation} framework that integrates instance-specific criteria generation with a criteria-aware scoring model. It achieves 83\% human alignment, significantly surpassing static-criteria baselines (65\%, 59\%). The effectiveness is further evidenced by its data curation capability-models trained with framework-filtered data outperform GPT-4o in writing.

% $\bullet$ We publicly release WritingBench, including its evaluation protocols, criteria generation tools with an integrated critic model, and writing-enhanced models, to foster further research. Available at: \url{https://github.com/X-PLUG/WritingBench}.

\section{Related Work}
\label{sec: related_work}

% The ongoing advancements in LLMs have prompted significant interest in developing benchmarks that effectively capture the capabilities and limitations of these models. Evaluating writing proficiency across diverse domains and requirements is crucial to understand their real-world applicability. This section reviews existing writing benchmarks, evaluation methods, and writing models, highlighting the unique contributions and limitations of each, while positioning our work within this landscape.

\subsection{Writing Benchmarks}

Existing writing benchmarks suffer from significant limitations in domain coverage and task granularity. For instance, EQ-Bench  (referring to its creative writing subset) encompasses templated queries for story-related tasks~\citep{eqbench}, while LongBench-Write incorporates length constraints in 120 queries~\citep{longwriter}; however, they both lack hierarchical domain taxonomies and multi-dimensional requirement specifications (e.g., style and format). Furthermore, most benchmarks rely on fixed instruction templates, short contexts, or materials predominantly from a single source~\citep{hellobench, nocha, pros-human-5}, rendering them insufficient for addressing the complexity of real-world needs. In contrast, our proposed benchmark fills these gaps by introducing 1,000 free-form queries distributed across 6 primary domains and 100 subdomains, with potential controls over style, format, and length, paired with inputs ranging from tens to thousands of words.

\begin{table}[t]
\centering
\small    
\caption{Comparison of existing writing benchmarks.}
\resizebox{1.0\textwidth}{!}{
\begin{tabular}{l|c|cc|ccc|cc|c|c} 
\toprule
\multirow{2.5}{*}{\textbf{Benchmark}} & \multirow{2.5}{*}{\textbf{Num}} & \multicolumn{2}{c|}{\textbf{Domains}} & \multicolumn{3}{c|}{\textbf{Requirement}}      & \multicolumn{2}{c|}{\textbf{Input Token}} & \multicolumn{1}{c|}{\multirow{2.5}{*}{\begin{tabular}[c]{@{}c@{}}\textbf{Free} \\\textbf{Query-Form}\end{tabular}}} & \multicolumn{1}{c}{\multirow{2.5}{*}{\begin{tabular}[c]{@{}c@{}}\textbf{Diverse} \\\textbf{Material-Source}\end{tabular}}}  \\ 
\cmidrule{3-9}
& & \textbf{Primary} & \textbf{Secondary} & \textbf{Style} & \textbf{Format} & \textbf{Length}  & \textbf{Avg}  & \textbf{Max}  & \multicolumn{1}{c|}{}  &   \\ 

\midrule

EQ-Bench & 241 & 1 & /  &   \textcolor{red}{\xmark}     &     \textcolor{red}{\xmark}   &   \textcolor{red}{\xmark} & 130& 213 &   \textcolor{red}{\xmark}  & / \\

LongBench-Write            & 120                      & 7    & /                     &  \textcolor{red}{\xmark}     &     \textcolor{red}{\xmark}   &     \textcolor{green}{\cmark}             & 87   & 684                         &                                                                                                \textcolor{green}{\cmark}     & /                                                                                \\
HelloBench                 & 647                      & 5    & 38                    &   \textcolor{red}{\xmark}     &     \textcolor{red}{\xmark}    & \textcolor{green}{\cmark}              & 1,210     & 7,766                      &                                                                                         \textcolor{red}{\xmark}            &      \textcolor{red}{\xmark}                                                                             \\

% LongProc &3616 &6&/ & \textcolor{green}{\cmark} &\textcolor{green}{\cmark} &\textcolor{red}{\xmark} &7320&90678&\textcolor{red}{\xmark}&\textcolor{red}{\xmark}\\

\textbf{WritingBench}        & 1,000                     & 6    & 100                   &    \textcolor{green}{\cmark}    & \textcolor{green}{\cmark}        &  \textcolor{green}{\cmark}                & 1,699 & 19,361                     &                                                                                                 \textcolor{green}{\cmark}    &   \textcolor{green}{\cmark}                                                                                \\
\bottomrule
\end{tabular}}
\label{tab:compare}
\vspace{-0.24cm}
\end{table}

\subsection{Evaluation Methods}

Using LLMs as evaluators has become a prevalent approach for evaluating the quality of generated responses. Typically, researchers pre-define a fixed set of evaluation dimensions applicable across all test instances~\citep{confederacy-human, pros-human-5}. For example, SuperCLUE~\citep{superclue} employs three dimensions, whereas LongBench-Write~\citep{longwriter} adopts six dimensions. HelloBench~\citep{hellobench} introduces task-specific dimensions, but the dimensions remain consistent across all queries of a given task. Although the LLM-as-a-judge approach enhances scalability, static evaluation dimensions often fail to accommodate the diversity of writing styles and specifications, especially when dealing with inputs with enriched materials. To address this limitation, recent work~\citep{fennec} trains a model to dynamically generate evaluation dimensions for individual queries. However, the dimensions remains confined to a small predefined set. In contrast, our query-dependent evaluation framework leverages LLMs to generate diverse and instance-specific criteria while fine-tuning a dedicated critic model to perform the evaluation.

% LLM-generated criteria with a fine-tuned critic model, enabling context-aware evaluation for complex writing tasks through three key innovations: (1) dynamic rubric adaptation based on query semantics, (2) multi-hop constraint verification, and (3) style-conditional scoring modules.

\subsection{Writing-Enhanced Models}

% Although existing LLMs showcase superior writing capabilities, researcher continuely to improve the writing upbound of LLMs. Recent models like Weaver~\citep{} and LongWriter~\citep{} have exhibited domain-specific strengths but face challenges in cross-domain adaptation. Weaver's 200B+ pretraining supports 4 writing domains ~\citep{} while Suri focuses on technical content generation~\citep{}. However, these systems show performance degradation when handling multi-constraint tasks requiring structural coherence across extended contexts~\citep{}. In this work, we train a writing-enhanced model and achieve competitive performance with chatgpt-4o-latest.

Although existing LLMs demonstrate exceptional writing capabilities, researchers strive to further enhance their overall writing proficiency. Recent models, such as Weaver~\citep{weaver} and Suri~\citep{suri}, have exhibited notable domain-specific strengths. For instance, Weaver benefits from over 200B parameter pretraining, supporting four distinct writing domains, while LongWriter~\citep{longwriter} specializes in length constraints. However, these models experience substantial performance degradation in cross-domain scenarios and multi-constraint tasks. In this work, leveraging the effectiveness of our evaluation framework, we introduce writing-enhanced models trained on high-quality data, achieving high performance across various tasks.

% Our benchmark's hierarchical organization enables systematic testing of contextual fidelity, particularly for models processing legal and academic documents exceeding 10k tokens~\citep{}.

% In summary, while previous efforts have laid important groundwork, existing benchmarks and evaluation methodologies often fall short of capturing the multifaceted requirements of comprehensive writing assessments. Our work introduces a structured and dynamic approach to these challenges, aiming to provide a robust tool for evaluating the evolving capabilities of LLMs in practical writing scenarios.

\section{WritingBench}
\label{sec: writingbench}

In this section, we will mainly introduce the four-stage human-AI collaborative construction process of WritingBench, and the query-dependent evaluation framework with its accompanied critic model for criteria-aware evaluation. Additionally, we train a writing-enhanced model to demonstrate the effectiveness of the framework's data curation capability.

\subsection{Benchmark Construction}
\label{ssec: benchmark_construction}

WritingBench is constructed following a delicate pipeline (see Figure~\ref{fig:Benchmark Construction}) that combines model-generated query refinement and human annotation, ensuring diversity and real-world relevance. The construction process is illustrated below.

\subsubsection{Model-Augmented Query Generation}
\label{sssec: model-augmented_query_generation}
This phase leverages LLMs to generate an initial set of writing queries, which are then enriched and diversified through systematic guidance, with material suggestions provided as needed.

%\begin{wrapfigure}[20]{r}{0.43\textwidth}
\begin{minipage}[t]{0.48\textwidth}
  \vspace{1.5em}
  \centering
    \raisebox{\dimexpr-\height+\baselineskip}[0pt][0pt]{%
    \includegraphics[width=\linewidth]{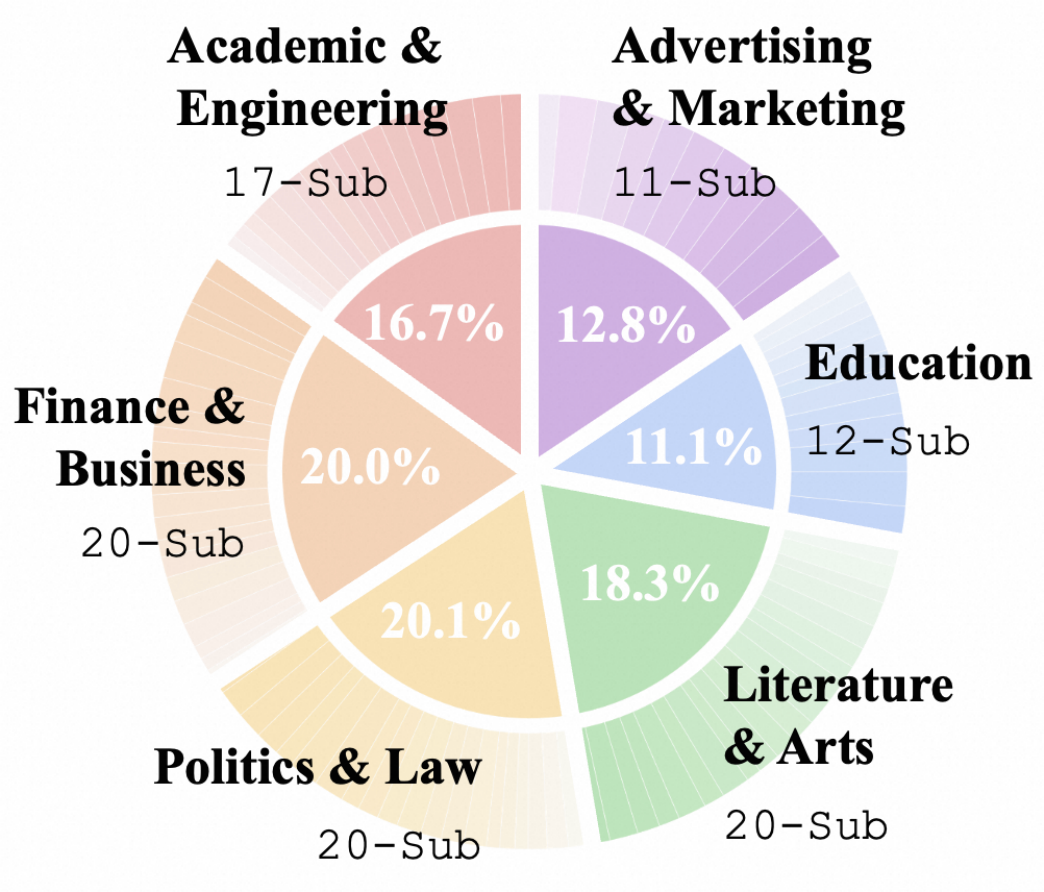}
  }
  \vspace*{-\baselineskip} % 补偿基线偏移
  \vspace{15em}
  \captionof{figure}{Domain categories in WritingBench. Inner represents the 6 primary domains and outer depicts 100 secondary subdomains (Sub = subdomains per category).}
  \label{fig:Domain}
\end{minipage}%\hspace{0.02\textwidth}% Add some space between the two 
\hfill
\begin{minipage}[t]{0.48\textwidth}
%\begin{wraptable}[18.5]{r}{0.48\textwidth}  % Adjust the width slightly to ensure the overall width fits
  %\vspace{-0.5em}
  %\footnotesize
  %\scriptsize
  \fontsize{8}{9}\selectfont
  \captionof{table}{Data statistics for WritingBench.}
  \centering
  \begin{tabular}{l|rrr} 
  \toprule
  \textbf{Category}      & \textbf{Num} & \begin{tabular}[c]{@{}c@{}}\textbf{Avg }\\\textbf{Token}\end{tabular} & \begin{tabular}[c]{@{}c@{}}\textbf{Max }\\\textbf{Token}\end{tabular}  \\

  \midrule
  \multicolumn{4}{c}{\cellcolor[gray]{0.9}\textit{Domain}}        \\ 
  \midrule
  Academic \& Engineering  & 167          & 1,944                                                                  & 15,534                                                                  \\
  Finance \& Business      & 210          & 1,867                                                                  & 19,361                                                                  \\
  Politics \& Law    & 201          & 2,363                                                                  & 18,317                                                                  \\
  Literature \& Arts       & 183          & 1,266                                                                  & 7,675                                                                   \\
  Education            & 111          & 1,345                                                                  & 10,737                                                                  \\
  Advertising \& Marketing & 128          & 984                                                                    & 6,504                                                                   \\ 

  \midrule
  \multicolumn{4}{c}{\cellcolor[gray]{0.9}\textit{Requirement}}                                                                                                                                         \\ 
  \midrule
  Style               &     442         &                                                                  1,728     &            19,361                                                                      \\
  Format               &       498       &    1,715                                                                &      15,534
  \\
  Length              &     222         &                                                                  1,362     &            14,097                                                       \\ 

  \midrule
  \multicolumn{4}{c}{\cellcolor[gray]{0.9}\textit{Length}}    \\                                                                                                               
  \midrule
0 - 1K     & 550 & 470   & 994     \\
1K - 3K   & 292 & 1,832 & 2,991   \\
3K - 5K   & 87  & 3,828 & 4,966   \\
Over 5K   & 71  & 8,053 & 19,361  \\
% \multicolumn{4}{c}{\cellcolor[gray]{0.9}\textit{Language}}    \\                                                                                                               
% \midrule
% Chinese         & 445          & 2,673                                                             & 19,361                                                                 \\
% English         & 555          & 916 & 3,750\\
  \bottomrule
  \end{tabular}
  \label{tab:Writing_benchmark_statistics}
  \vspace{0.3cm}
\end{minipage}
%\vspace{-0.5em}

% \subsubsection{Domain Taxonomy}
% \label{sssec: domain_taxonomy}

\vspace{1em}    
\noindent\textbf{Phase 1 - Initial Query Generation:}
We begin by constructing a two-tiered domain pool grounded in real-world writing scenarios, consisting of 6 primary domains and 100 subdomains (see Figure~\ref{fig:Domain} and Appendix~\ref{app: Benchmark Statistics} for detailed domain statistics and descriptions; the construction process of domain system is described in Appendix~\ref{app: domain_taxonomy_construction}). These domains are designed to capture both traditional and emerging user needs for AI-assisted writing, categorized by topic and functionality into 6 primary domains: Academic \& Engineering, Finance \& Business, Politics \& Law, Literature \& Art, Education, and Advertising \& Marketing. 
%These domains are designed to capture both traditional and emerging user needs for AI-assisted writing, encompassing categories such as Academic \& Engineering, Finance \& Business, Politics \& Law, Literature \& Art, Education, Advertising \& Marketing. 
Leveraging the primary domain and subdomain tags, we prompt two different LLMs (GPT-4o and Claude-3.5-Sonnet~\cite{claude-3.5})% to generate initial writing queries 
to produce an extensive pool of initial writing query drafts that simulate realistic user requests to maximize diversity (see Appendix~\ref{app:InitialQueryGenerationPrompt}).

\vspace{0.5em}
\noindent\textbf{Phase 2 - Query Diversification:}
To enhance the diversity and practical applicability of queries while addressing real-world needs, we design a set of query diversification strategies (see Appendix~\ref{app:GuidancePool}), inspired by \cite{wizardlm}. These strategies are divided into three core requirements and three auxiliary dimensions. The core requirements focus on fundamental aspects of writing: (1)~Stylistic adjustments (e.g., "Use a friendly and simple tone that kids can easily understand"), (2)~Format specifications (e.g., "Follow the IEEE conference template"), and (3)~Length constraints (e.g., "Generate a 500-word executive summary"), which will be evaluated through specialized assessments. The auxiliary dimensions address additional considerations: (4)~Personalization (e.g., "From the viewpoint of an educator with two decades of experience"), (5)~Content specificity (e.g., "Detail the 2023 Q3 financial metrics"), and (6)~Expression (e.g., "Modify the query expression to be shorter"). While refining the queries, the LLM simultaneously provides necessary material suggestions (e.g., requesting financial reports as input for market analysis queries) (see Appendix~\ref{app:QueryRefinePrompt}). This structured approach ensures that queries are both diverse and practical.

\begin{figure*}[!t]
  \centering
  \includegraphics[width=1\textwidth]{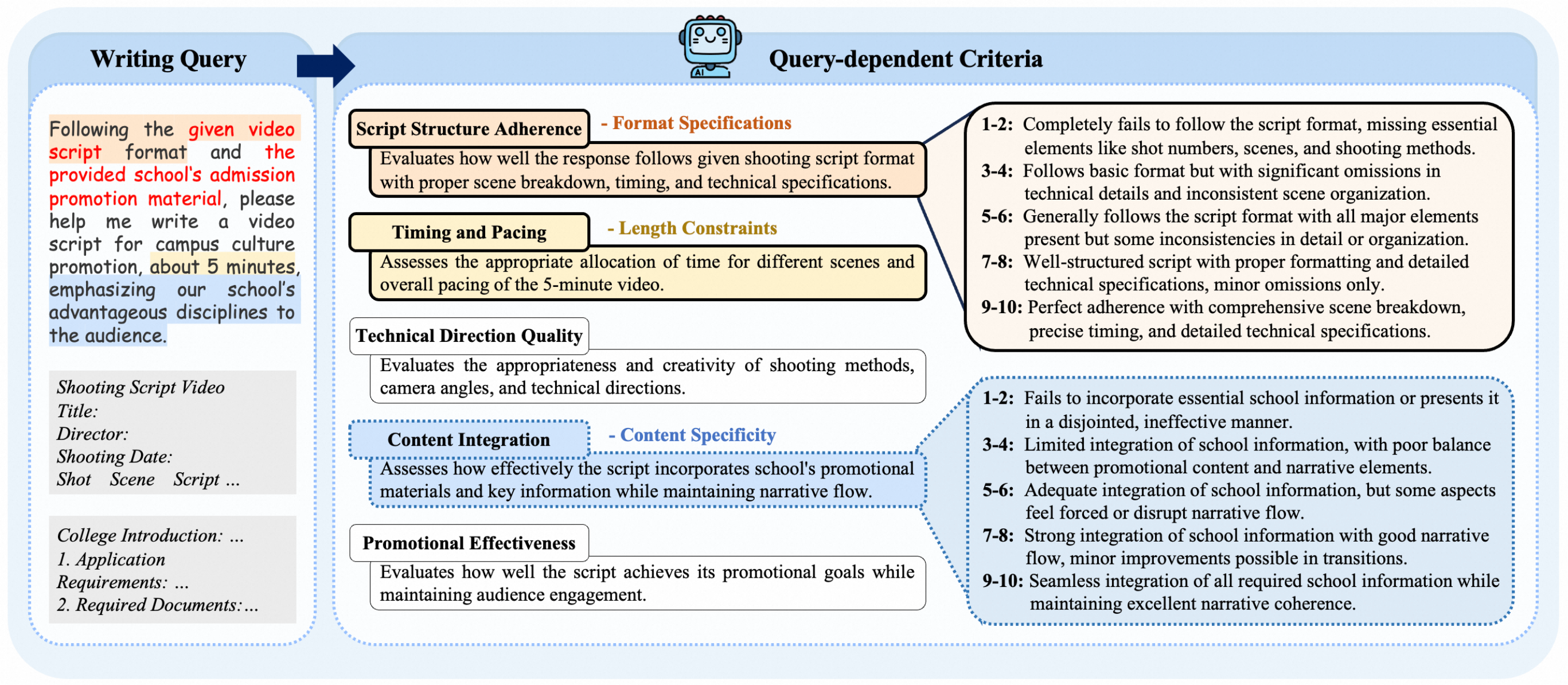}
  \caption{Example of dynamically generating criteria for a writing query in WritingBench. Different background colors represent various types of requirements.}
  \label{fig:criteria_example}
  % \vspace{-0.35cm}
\end{figure*}

\subsubsection{Human-in-the-Loop Refinement}
\label{sssec: human-in-the-loop_refinement}
The above outputs serve solely to provide a large, linguistically diverse pool of drafts. Human experts then verify these queries and supplement the required materials, ensuring alignment with real-world applications and avoiding harmful content.

\vspace{0.5em}
\noindent\textbf{Phase 1 - Material Collection:}
At this stage, we engage 30 trained annotators, compensated at \$18 per hour and possessing specialized expertise proven through domain-knowledge tests. Their role is to collect open-source reference materials needed for the writing tasks (e.g., public financial statements or legal templates). Annotators may either adopt the sources suggested by the LLM-generated material requirements or, when more appropriate, independently identify alternative documents. Each query can be paired with multiple material sources, whereas in some cases no external material is required. To minimize errors caused by parsing documents in varied formats and to eliminate noise from irrelevant content such as web advertisements, the annotators carefully extract and verify only the most pertinent text segments.

\vspace{0.5em}
\noindent\textbf{Phase 2 - Expert Optimization:}
% Subsequently, we invite six experts to perform data screening. All experts have experience with the use of LLMs or are professionals in the related industry. The experts performed dual filtering: (1)~query adaptation: rewrite ambiguous or unrealistic queries to better align with materials and practical scenarios (e.g., adjusting a legal opinion query to reference specific clauses from provided statutes). (2)~material pruning: remove redundant or irrelevant content from collected materials and conduct a thorough review to ensure the data is free from harmful content, thereby maintaining a focused context for writing tasks.
Next, five experts with LLM experience or relevant industry expertise conduct data screening. The experts first exclude low-quality data that are unrealistic, overly generic, or potentially harmful. For the remaining entries, a rigorous two-stage review is employed: (1)~query adaptation: ambiguous expressions are revised to better align with the provided materials and practical scenarios (e.g., adjusting a legal opinion query to reference specific clauses from the supplied statutes), (2)~material pruning: redundant or irrelevant information is eliminated from the accompanying materials, ensuring that the context for writing tasks remains focused, relevant, and devoid of harmful content. Furthermore, experts are encouraged to create original, high-quality queries that are cross-validated by their peers, to further enhance diversity and realism.

% Finally, we obtain our WritingBench, which consists of 1,239 queries organized through a two-tiered taxonomy, as illustrated in Figure~\ref{fig:Domain}. Compared with existing writing benchmarks shown in Table~\ref{tab:compare}, WritingBench demonstrates comprehensive advantages over existing ones in instance number, domain diversity, requirement coverage, and length expansion. The detailed statistics of WritingBench are shown in Table~\ref{tab:Writing_benchmark_statistics}.

Finally, we construct WritingBench, a benchmark comprising 1,000 queries categorized using a two-tiered taxonomy, including 445 queries in Chinese and 555 queries in English. In comparison to existing writing benchmarks summarized in Table~\ref{tab:compare}, WritingBench exhibits notable advantages in terms of the number of instances, domain diversity, requirement coverage, and variability in input lengths. The detailed statistical distribution of WritingBench is shown in Table~\ref{tab:Writing_benchmark_statistics}.

% Finally, each query was associated with five evaluation criteria (see §3.3) to assess model outputs’ adherence to multifaceted requirements. This hybrid approach balanced scalability (via LLM augmentation) with fidelity (via human expertise), resulting in a benchmark that captures the complexity of real-world writing tasks across diverse domains.

\subsection{Query-Dependent Evaluation Framework}
\label{ssec: evaluation_metric}

Traditional LLM-as-a-judge evaluations typically rely on fixed evaluation criteria derived from general writing assessment conventions~\citep{storm, longwriter, hellobench}. However, such static criteria exhibit three critical limitations: (1)~Domain exhaustiveness: fixed criteria are inadequate in adapting to specialized domains, failing to address unique characteristics found in areas like technical documentation or creative writing; (2)~Requirement specificity: such criteria lack the flexibility to encompass specific requirements related to style, format, or length control; and (3)~Material dependency: they are insufficient to verify whether responses appropriately utilize the provided reference materials, such as incorporating data points or maintaining narrative continuity. To address these challenges, we propose a query-dependent evaluation framework that enables dynamic adaptation to diverse writing scenarios. This approach comprises two phases:

\vspace{0.5em}
\noindent\textbf{Phase 1: Dynamic Criteria Generation:}
% Given a query $q$ in the WritingBench, we ask the LLM to automatically derive 5 evaluation criteria $C_q=\{C_1,\ldots,C_5\}$ using well-designed instruction to provide structural guidance for criteria specification (as shown in Appendix~\ref{}).
% \begin{equation}
%     C_q = \{C_1,\ldots,C_5\} = f_{\text{criteria}}\big(q(m), P_{\text{template}}\big)
% \end{equation}
% where $q(m)$ denotes material-augmented query integration, and $P_{\text{template}}$ 
% Each criterion includes a succinct name describing the criterion, an extended description detailing the evaluation focus, and detailed rubrics. The scoring rubrics define granular quality levels for each criteria category.
As illustrated in Figure~\ref{fig:criteria_example}, given a query $q$ in the WritingBench, the LLM is prompted to automatically generate a set of five evaluation criteria, $C_q = \{c_1, \ldots, c_5\}$ (opting for five criteria reflects a common approach observed in many contemporary evaluation contexts~\citep{storm, mops, pros-human-5}), using a carefully designed instruction to ensure structural guidance during criteria specification (see Appendix~\ref{app:criteria generation prompt}). We utilize Claude-3.7 for generation, as it demonstrates superior diversity and comprehensiveness in criteria generation compared to models such as GPT-4o(see Appendix~\ref{app:criteria_comparison} for details). Each criterion comprises three components: a concise name summarizing the criterion, an extended description elaborating on the evaluation focus, and detailed scoring rubrics, which provide fine-grained quality levels for the respective evaluation dimensions. The generated criteria are further reviewed by human annotators to confirm their reasonableness and ensure no harmful content.

\begin{table*}[!t]
\centering
\caption{WritingBench performance of LLMs across 6 domains and 3 core requirements evaluated with our critic model (scale: 1-10). The standard deviation is computed over 3 samples. Domains include: (D1) Academic \& Engineering, (D2) Finance \& Business, (D3) Politics \& Law, (D4) Literature \& Art, (D5) Education, and (D6) Advertising \& Marketing. The writing requirements assessed are: (R1) Style, (R2) Format, and (R3) Length. Here, ``C" indicates category-specific scores. The latest results are available on the online leaderboard.}
\footnotesize
\resizebox{1.0\textwidth}{!}{
\begin{tabular}{l|c|cc|cccccc|clclcl}
\toprule
\multirow{2.5}{*}{\textbf{Models}} & \multirow{2.5}{*}{\textbf{Overall}} & \multicolumn{2}{c|}{\textbf{Languages}} & \multicolumn{6}{c|}{\textbf{Domains}} & \multicolumn{6}{c}{\textbf{Requirements}} \\
\cmidrule(l){3-16}
& & \textbf{ZH} & \textbf{EN} & \textbf{D1} & \textbf{D2} & \textbf{D3} & \textbf{D4} & \textbf{D5} & \textbf{D6} & \textbf{R1} & \multicolumn{1}{c|}{\textbf{C}} & \textbf{R2} & \multicolumn{1}{c|}{\textbf{C}} & \textbf{R3} & \multicolumn{1}{c}{\textbf{C}} \\

\midrule
\multicolumn{16}{l}{\cellcolor[gray]{0.9}\textit{Proprietary LLMs}} \\
\midrule
Claude-3.7-thinking & \textbf{7.91}$_{\pm 0.111}$ & 7.9 & 7.9 & 7.9 & 7.8 & 7.8 & 8.0 & 8.0 & 8.1 & 7.9 & 8.7 & 8.0 & 8.4 & 8.0 & 8.1 \\
Claude-3.7 & 7.85$_{\pm 0.110}$ & 7.9 & 7.8 & 7.8 & 7.8 & 7.7 & 7.9 & 8.0 & 8.1 & 7.9 & 8.6 & 7.9 & 8.3 & 8.0 & 8.1 \\
Qwen-Max & 7.16$_{\pm 0.041}$ & 7.2 & 7.1 & 7.1 & 6.9 & 7.0 & 7.3 & 7.4 & 7.5 & 7.2 & 8.3 & 7.3 & 7.8 & 7.2 & 7.5 \\
o1-Preview & 6.89$_{\pm 0.039}$ & 6.8 & 7.0 & 6.9 & 6.8 & 6.7 & 7.0 & 7.1 & 7.2 & 6.9 & 8.0 & 7.0 & 7.5 & 7.1 & 7.3 \\
GPT-4o & 6.81$_{\pm 0.028}$ & 6.9 & 6.7 & 6.8 & 6.6 & 6.7 & 6.8 & 7.0 & 7.1 & 6.9 & 8.0 & 7.0 & 7.5 & 6.8 & 6.8 \\
Gemini-1.5-Pro & 6.21$_{\pm 0.018}$ & 6.2 & 6.2 & 6.2 & 5.8 & 6.0 & 6.4 & 6.6 & 6.7 & 6.2 & 7.2 & 6.4 & 7.1 & 6.4 & 6.0 \\

\midrule
\multicolumn{16}{l}{\cellcolor[gray]{0.9}\textit{Open-source LLMs}} \\
\midrule
Deepseek-R1 & 7.70$_{\pm 0.053}$ & 8.0 & 7.5 & 7.6 & 7.4 & 7.6 & 7.8 & 7.8 & 8.1 & 7.7 & 8.4 & 7.9 & 8.3 & 7.7 & 7.5 \\
Deepseek-V3 & 6.35$_{\pm 0.022}$ & 6.3 & 6.4 & 6.4 & 6.1 & 6.2 & 6.3 & 6.6 & 6.8 & 6.4 & 7.6 & 6.5 & 7.1 & 6.5 & 6.4 \\
Mistral-Large-Instruct & 6.00$_{\pm 0.076}$ & 5.9 & 6.1 & 6.2 & 5.9 & 5.9 & 5.7 & 6.4 & 6.4 & 6.1 & 7.3 & 6.1 & 6.5 & 6.0 & 6.0 \\
Qwen-2.5-72B-Instruct & 6.40$_{\pm 0.061}$ & 6.4 & 6.4 & 6.6 & 6.2 & 6.4 & 6.2 & 6.7 & 6.6 & 6.5 & 7.7 & 6.5 & 6.9 & 6.5 & 6.5 \\
Qwen-2.5-7B-Instruct & 5.64$_{\pm 0.083}$ & 5.5 & 5.8 & 5.9 & 5.6 & 5.6 & 5.1 & 6.1 & 5.9 & 5.7 & 7.0 & 5.7 & 6.1 & 5.6 & 5.6 \\
Llama-3.3-70B-Instruct & 5.05$_{\pm 0.011}$ & 4.5 & 5.5 & 5.1 & 4.9 & 4.8 & 4.8 & 5.3 & 5.6 & 5.0 & 5.0 & 5.1 & 5.9 & 5.1 & 5.0 \\
Llama-3.1-8B-Instruct & 4.42$_{\pm 0.004}$ & 3.7 & 5.0 & 4.1 & 4.4 & 4.0 & 4.1 & 4.7 & 5.0 & 4.4 & 4.4 & 4.5 & 5.3 & 4.4 & 4.3 \\

\midrule
\multicolumn{16}{l}{\cellcolor[gray]{0.9}\textit{Capability-enhanced LLMs}} \\
\midrule
Suri & 3.20$_{\pm 0.042}$ & 2.5 & 3.8 & 3.6 & 3.5 & 3.0 & 2.5 & 3.2 & 3.6 & 3.2 & 3.7 & 3.1 & 3.2 & 3.0 & 3.0 \\
LongWriter & 6.27$_{\pm 0.081}$ & 6.2 & 6.4 & 6.4 & 6.4 & 6.3 & 6.0 & 6.5 & 6.0 & 6.3 & 7.4 & 6.3 & 6.7 & 6.3 & 6.8 \\
Qwen-2.5-7B-filtered & 7.44$_{\pm 0.058}$ & 7.7 & 7.2 & 7.4 & 7.2 & 7.5 & 7.3 & 7.7 & 7.7 & 7.5 & 8.4 & 7.6 & 8.1 & 7.4 & 7.2 \\
Llama-3.1-8B-filtered & 7.39$_{\pm 0.045}$ & 7.5 & 7.3 & 7.4 & 7.2 & 7.3 & 7.3 & 7.5 & 7.8 & 7.4 & 8.3 & 7.5 & 8.0 & 7.4 & 7.1 \\

\bottomrule
\end{tabular}}
\label{tab:main table}
% \vspace{-0.46cm}
\end{table*}

We identify three commonly encountered writing requirements: style, format, and length from real-world writing scenarios (frequency analysis detailed in Appendix~\ref{app: domain_taxonomy_construction}). For each requirement, we create two evaluation subsets. Two annotators review each query to classify each criteria in $C_q$ into style, format, length, or none. Based on their annotations, we perform final verification and define the subsets as follows: (1)~The first requirement-involved subset includes all queries involving a specific requirement and their entire set of criteria. For example, if any criterion in $C_q$ relates to style, the query and its entire criteria set $C_q$ belong to this subset. Scores for all five criteria are calculated, corresponding to the R1/R2/R3 columns in Table~\ref{tab:main table}. (2)~The second category-specific subset includes only the criteria related to the specific requirement. For instance, if $c_1$ and $c_4$ in $C_q$ are format-related, then the query and the subset ${c_1, c_4}$ belong to this subset. Only the scores for ${c_1, c_4}$ are calculated, corresponding to the C columns. The first subset serves as an overall evaluation of writing quality for a specific requirement, while the second subset provides more targeted insights into performance on that particular capability.

\vspace{0.5em}
\noindent\textbf{Phase 2 - Rubric-based Scoring:}
% For each criterion $c_i \in C_q$, we further ask the LLMs to assign an independent 10-point scale evaluation on a given response $r$.
% \begin{equation}
%     \text{Score}_i(q) = f_{\text{score}}\big(q(m), r, C_i\big) \quad \forall i \in \{1,\ldots,5\}
% \end{equation}
% When scoring, the model is required to provide a score and detailed rating reasons. The final score is aggregated by averaging the scores of different dimensions. Detailed prompts for this process can be found in Appendix~\ref{}.
For each criterion $c_i \in C_q$, the evaluator independently assigns a score on a 10-point scale to a response $r$, along with a justification. The final overall score is computed by averaging scores across all criteria. Scoring prompt is provided in Appendix~\ref{app:rubric-based scoring prompt}.

To alleviate the computational overhead with LLM-based evaluation, we develop a dedicated critic model, $\mathcal{M}$, designed to implement our rubric-based scoring framework. Specifically, this model performs the mapping $\mathcal{M}_c: (q, r, c_i) \mapsto [1,10] \times \mathcal{J}$, where the output consists of a numerical score and corresponding justification text, $\mathcal{J}$, both in accordance with the predefined evaluation rubric. 
The critic model is fine-tuned on a dataset of 155K instances scored by Claude-3.7, which demonstrates higher human alignment consistency (as discussed in Section~\ref{ssec:consistency_exp}). When building the training data of scoring prompts, the queries and criteria are drawn from WritingBench. The response portion is generated using approximately 40 different models, including those evaluated in Section~\ref{ssec:main}, as well as additional models of varying types and sizes, such as Claude-3.5-Haiku~\cite{claude-3.5} and Llama-3.2-1B-Instruct~\cite{llama-3.2}. Finally, after removing instances where scoring failed, approximately 155K samples are constructed for training. The experiments described in Section~\ref{ssec:consistency_exp} confirm the consistency of the critic model.

% We fine-tune the critic model using a dataset of 50K instances, which are collected using LLMs from our experiments covering various queries, criteria, and model responses to ensure the robustness in evaluation. The training details are shown in Appendix~\ref{ssec:training} and experiments in Section~\ref{} demonstrate the consistency of the critic model.

% The criteria within this dataset are generated through our dynamic criteria generation process, ensuring a degree of  Experiments in~\ref{} demonstrate that the critic model achieves a commendable level of consistency with human evaluators.

\subsection{Evaluation-Guided Data Curation for Writing Enhancement}
\label{ssec: writing_model}

The query-dependent evaluation framework enables systematic data curation across diverse writing tasks through two-phase filtering. To validate its effectiveness in data curation, we conduct SFT experiments using constructed datasets. We utilize the secondary domain taxonomy of WritingBench and follow the first two steps outlined in Section~\ref{sssec: model-augmented_query_generation} (removing the material generation suggestion part from the prompt in Appendix~\ref{app:QueryRefinePrompt}, requiring only the refined query from the model) to let LLMs generate writing queries. To enrich query diversity, we employ two models, GPT-4o and Claude-3.5-Sonnet, to each generate 60 queries in both Chinese and English for every subdomain, totaling 24K queries. Responses are uniformly generated using an advanced model, Deepseek-R1. 

Subsequently, we apply the query-dependent evaluation metric with our critic model described in Section~\ref{ssec: evaluation_metric} to select top-50\% samples per subdomain, resulting in 12K high-quality data samples. Fine-tuning experiments are conducted using the Llama-3.1-8B-Instruct and Qwen-2.5-7B-Instruct models. Both models (Qwen-2.5-7B-filtered and Llama-3.1-8B-filtered) demonstrated significant performance improvements over their base versions and even outperformed larger models such as Llama-3.3-70B-Instruct and Qwen-2.5-72B-Instruct in our experiments on two benchmarks, confirming the framework's effectiveness in identifying high-quality writing samples.

\section{Experiment}
\label{sec: experiment}

\begin{figure*}[t]
  \centering
  \includegraphics[width=1\textwidth]{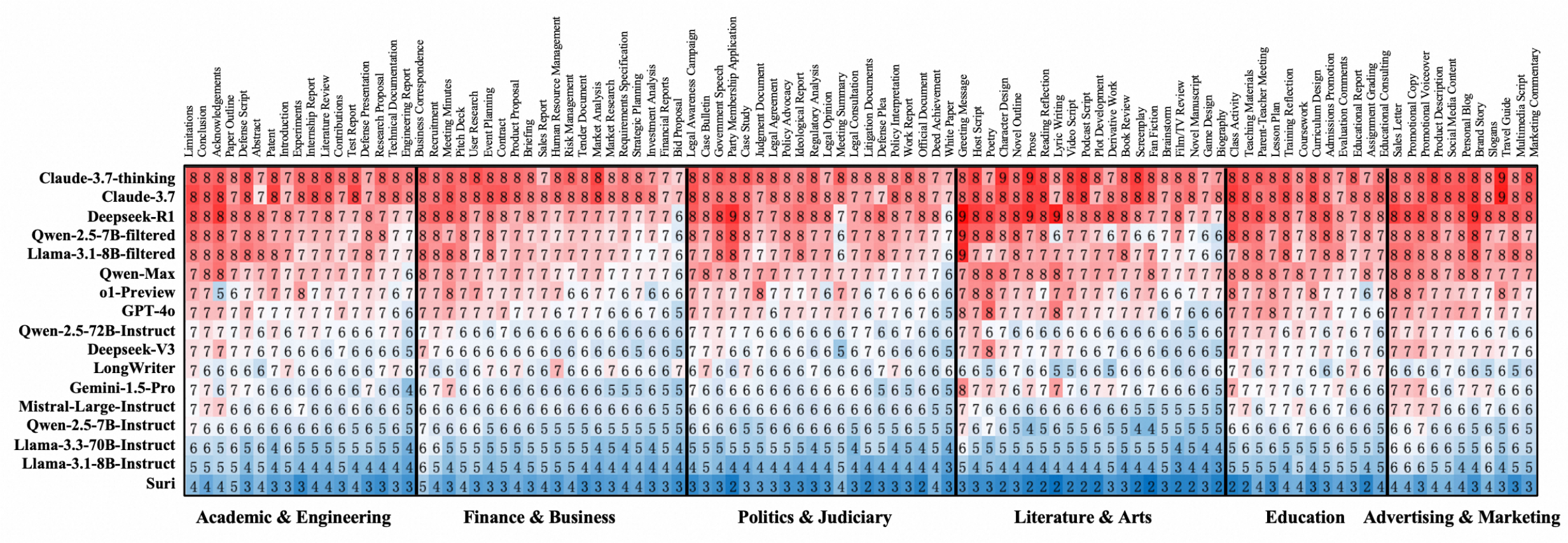}
  \caption{Score heatmap of models across 100 subdomains in WritingBench. The figure shows the average score per subdomain for each model. Red indicates higher score and blue indicates lower.}
  \label{fig:domain2}
  \vspace{-0.17cm}
\end{figure*}

\subsection{Experiment Settings}
\label{ssec: experiment_settings}

This section outlines the settings employed in our experiments using the WritingBench framework. The evaluation protocol and the model training configurations are as follows: % Our approach ensures accurate and consistent performance assessment through advanced AI-assisted scoring and human evaluation. % We detail the evaluation protocol, and the model training configurations, forming a rigorous experimental setup (see Appendex~\ref{app:experiment settings}). These components are meticulously outlined to facilitate reproducibility and transparency, enabling other researchers to replicate and extend our findings.

% \section{Experiment Settings}
% \label{app:experiment settings}

% \subsection{Evaluation Protocol}
% \vspace{0.5em}

\noindent\textbf{Evaluation Protocol:} %\\
Uniform generation settings are applied to model responses on WritingBench, with maximum sequence lengths capped at 16,000 tokens (or platform maximum if lower). We configure the generation process with a temperature of 0.7, a top-k value of 20, and top-p of 0.8, ensuring a balance of creativity and coherence. 
For both the critic model and LLM scoring, we use consistent configurations. The temperature is set to 1.0, in line with the Arena-Hard-Auto framework~\cite{arena-hard}. We apply a top-p value of 0.95 without any top-k filtering, and the maximum length is set at 2,048 tokens to ensure efficient and reliable output generation.

% Beyond general quality assessment, we implement specialized evaluations for style, format, and length subsets. This is reflected in column R of Table~\ref{tab:main table}, which represents the average score across five criteria. Column C represents the specialized capability score, averaged by criteria corresponding to each capability.

% \subsection{Model Training} 
% \vspace{0.5em}
\noindent\textbf{Model Training:} % \\
% \label{ssec:training}
The critic model is fine-tuned on the Qwen-2.5-7B-Instruct base model, using the AdamW optimizer with a 7e-6 learning rate. It is trained on 155K SFT data for 3 epochs across 16xA100 GPUs (batch size 64, 4-step accumulation). The writing models, trained on both Qwen-2.5-7B-Instruct and Llama-3.1-8B-Instruct for 5 epochs on 32xA100 GPUs, achieving a total batch size of 128 with 4-step gradient accumulation.

\begin{table}[t]
\centering
\begin{minipage}[!t]{0.48\textwidth}
  \centering
  \small
  \captionof{table}{Comparison of human consistency scores across different criteria generation methods. Claude corresponds to Claude-3.7. Critic refers to the critic model.}
  \begin{tabular}{ll|c}
    \toprule
    \textbf{Evaluation Metric} & \textbf{Judge} & \textbf{Score}  \\
    \midrule
    Static Global & GPT-4o & 69\%  \\
    Static Domain-Specific & GPT-4o & 40\%  \\
    Dynamic Query-Dependent & GPT-4o & 79\% \\   
    \midrule
    Static Global & Claude & 67\%  \\
    Static Domain-Specific & Claude & 58\%  \\
    \textbf{Dynamic Query-Dependent} & Claude & 87\% \\  
    \midrule
    \textbf{Dynamic Query-Dependent} & \textbf{Critic} & 84\% \\       
    \bottomrule
  \end{tabular}
  \label{tab:agreement_scores}
\end{minipage}%
\hspace{0.02\textwidth} % 增加间距
\begin{minipage}[!t]{0.48\textwidth}
  \centering
  \small
  \captionof{table}{Performance of our writing models on two benchmarks. `-filtered' indicates models trained with filtered data, while `-all' uses the full dataset. Eval2 refers to LongBench-Write~\cite{longwriter}.}
  \begin{tabular}{l|cc}
    \toprule
    \textbf{Models} & \textbf{WritingBench} & \textbf{Eval2}  \\
    \midrule
    Deepseek-R1 & 7.70 & 4.79  \\
    Qwen-2.5-7B-Instruct & 5.64 & 4.39 \\
    Llama-3.1-8B-Instruct & 4.42 & 3.12 \\ 
    \midrule
    Qwen-2.5-7B-all & 7.36 & 4.69  \\
    Qwen-2.5-7B-filtered & 7.44 & 4.70  \\
    Llama-3.1-8B-all & 7.34 & 4.65 \\  
    Llama-3.1-8B-filtered & 7.39 & 4.65 \\   
    \bottomrule
  \end{tabular}
  \label{tab:writing_model}
\end{minipage}
\end{table}

\subsection{Comparison between LLMs}
\label{ssec:main}

We evaluate 17 LLMs: GPT-4o\footnote{In this paper, we specifically use GPT-4o, version gpt-4o-2025-01-29.}~\citep{4ol}, o1-Preview~\citep{o1}, Claude-3.7 and its thinking version Claude-3.7-thinking~\citep{claude}, Gemini-1.5-Pro~\citep{gemini}, Qwen-Max~\citep{qwen-max}, Deepseek-R1~\citep{r1}, Deepseek-V3~\citep{deepseek-v3}, Mistral-Large-Instruct~\citep{mistral}, Qwen-2.5-72B-Instruct and Qwen-2.5-7B-Instruct~\citep{qwen2.5}, Llama-3.3-70B-Instruct and Llama-3.1-8B-Instruct~\citep{llama}, Suri~\citep{suri}, LongWriter~\citep{longwriter}, and our writing model, Qwen-2.5-7B-filtered (fine-tuned on Qwen-2.5-7B-Instruct) and Llama-3.1-8B-filtered (fine-tuned on Llama-3.1-8B-Instruct), on WritingBench. For models capable of reasoning, the reasoning content is excluded from the evaluated response. Each query is assessed by 5 instance-specific criteria using a 10-point scale scoring by our critic model. The overall scores of the models are presented in Table~\ref{tab:main table}, along with their performance across two languages and various domain and requirement subsets. Detailed variations are further revealed through subdomain-specific subcategory heatmap in Figure~\ref{fig:domain2}.

\noindent\textbf{Key Insights from Domain Scores:} 
The Education (D5) and Advertising \& Marketing (D6) domains always yield substantial performance across models.  In contrast, the Academic \& Engineering (D1) and Finance \& Business (D2) domains present greater challenges, as these tasks inherently require more sophisticated information processing and integration capabilities. Our evaluation across 100 subdomains further identifies persistent difficulties in niche areas like writing bid proposals, financial reports, and white papers, where models generally achieve lower scores. These tasks demand a higher level of knowledge, long-text generation capabilities, and adherence to contextual consistency, pinpointing areas for further enhancement.

The Literature \& Art (D4) domain exhibits notable performance variance among models. Reasoning-capable architectures such as Claude-3.7-thinking, Deepseek-R1, and o1-Preview outperform their non-reasoning counterparts, indicating the potential of CoT techniques in processing narrative and creative content. To further validate CoT’s efficacy in creative writing, we employ the 12k SFT dataset described in Section~\ref{ssec: writing_model} to fine-tune Qwen-32b-Instruct using both CoT-integrated and non-CoT approaches. Evaluations are conducted on the Literature \& Art (D4) subset of WritingBench and EQ-Bench~\cite{eqbench}, a specialized benchmark designed for creative writing tasks. The results reveal that the reasoning model consistently surpass both the baseline Qwen-32b-Instruct and its non-reasoning variants, with detailed results provided in Appendix~\ref{app:creative}.

\noindent\textbf{Key Insights from Requirement Scores:} 
Most models perform well in the style dimension, followed by format, with length being the weakest. We observe that advanced models often achieve higher specialized scores (C column) for the three common requirements compared to their overall scores (R column). Criteria outside these specialized sets tend to emphasize content-related aspects, such as integration with source materials and writing depth, underscoring the need to improve content quality.

Length requirements remain particularly difficult, especially in section-specific constraints (see Appendix~\ref{app:length bad case} for examples) and extended text generation. Additionally, we evaluate the performance of LLMs across varying input and output lengths (see Appendix~\ref{app:lenth_exp} for details). Advanced models generally sustain consistent performance across varying input lengths, leveraging their strong long-context comprehension abilities. However, regarding output length, most models demonstrate inherent limitations, typically capping their responses at around 3,000 tokens. In contrast, Claude-3.7 and its reasoning-enhanced version, as well as Qwen-Max, stand out for their capacity to generate extended outputs effectively. These results emphasize the critical need to improve long-output generation and refine length optimization in writing-related tasks.

The overall analysis of WritingBench experiment highlights:  (1)~Claude-3.7-thinking consistently leads across both domain and requirement dimensions, followed by its non-reasoning variant, showcasing versatility and strong language capabilities; (2)~A significant performance variance is observed within creative content domains, where models incorporating CoT mechanisms surpass those without, demonstrating CoT's potential in LLM writing; (3)~Cross-lingual inconsistencies in models like Deepseek-R1 and Llama-3.3-70B-Instruct suggest limitations in multilingual knowledge alignment. Furthermore, detailed analysis on subdomains reveals persistent difficulties in knowledge-intensive tasks. Ablation experiments on output length emphasize that generating long-form outputs remains an obstacle for current models. This analysis not only benchmarks the existing capabilities of these models but also underscores specific areas needing improvement for future development. For the most up-to-date results, please refer to the online leaderboard\footnote{For clearer display, online leaderboard scores use the same calculations but are rescaled to a 100-point scale.}.

\subsection{Human Consistency}
\label{ssec:consistency_exp}

To validate the alignment between automated evaluation and human judgment, we conduct a human evaluation study involving 300 queries (constructed in the same pipeline as described in Section~\ref{ssec: benchmark_construction} but not included in WritingBench). Five professionally trained annotators with linguistic backgrounds perform pairwise comparisons of model responses. For each query, two responses are generated by two randomly selected models (drawn from the same set of 17 models used for evaluation in Section~\ref{ssec:main}). The triplet <Query, Response\_A, Response\_B> is presented on the same page to enable direct comparison. Annotators are instructed to carefully read the query and select their preferred response or declare equivalence (A/B/Tie), with no criteria information provided to ensure unbiased decision-making (detailed instructions and the annotation interface are provided in Appendix~\ref{app:human_consistency}).

We then compare these human preferences with the scores assigned by our critic model to the responses — if the critic model rate Response\_A higher and the annotators also prefer Response\_A, it is counted as alignment. To further evaluate the effectiveness of our dynamic query-dependent criteria, we compare it against two baselines: static globally uniform criteria and static domain-specific customized criteria (designed by domain experts). This comparison was conducted using two LLM-based judges, GPT-4o and Claude-3.7. Both models score the responses using the rubric-based scoring method described in Section~\ref{ssec: evaluation_metric}, using scoring prompts provided in Appendix~\ref{app:rubric-based scoring prompt}.

As shown in Table~\ref{tab:agreement_scores}, our dynamic query-dependent criteria achieve superior human alignment compared to static, both globally uniform or domain-specific customized criteria. Notably, domain-specific criteria underperform despite customization, since our queries involve highly diverse tasks and varied sources. These findings confirm that context-sensitive query-dependent evaluation better captures real-world writing complexity compared to conventional static approaches. Furthermore, the critic model attains 84\% agreement, confirming its practical viability.

\subsection{Ablation of Data Curation for Writing-Enhanced Models} 
\label{sec:ablation}

To validate the data curation capabilities of the query-dependent evaluation framework described in Section~\ref{ssec: evaluation_metric}, we conduct fine-tuning experiments on two datasets: the initial 24K dataset constructed in Section~\ref{ssec: writing_model} and the 12K subset curated using the query-dependent evaluation framework. We experiment with two models of different architectures, Llama-3.1-8B-Instruct and Qwen-2.5-7B-Instruct, and evaluate their performance on two benchmarks: WritingBench and LongBench-Write, a general-purpose writing benchmark (following the quality evaluation settings outlined in~\citep{longwriter}). 

As shown in Table~\ref{tab:writing_model}, both models trained on the filtered 12K dataset demonstrate significant performance improvements over their previous versions. Notably, they outperform models trained on the full 24K dataset and even approach the capabilities of advanced models. These results validate the robustness of our query-dependent evaluation strategy and highlight the effectiveness of our critic model in curating high-quality writing samples. This integrated approach enables smaller models to compete with, and in some cases surpass, larger models across a wide range of writing tasks.

\section{Conclusion}
\label{sec: conclusion}

In this paper, we introduce WritingBench, a comprehensive benchmark designed to evaluate LLMs in generative writing across diverse domains. It includes 1,000 queries spanning 6 primary domains and 100 subdomains, providing evaluation dimensions for style, format, and length requirements. Our query-dependent evaluation framework, supported by a critic model, achieves high human alignment. Evaluation efficiency is further demonstrated by compact models trained on curated data, outperforming GPT-4o in writing. By making WritingBench and its resources publicly available, we aim to foster further research and advancements in LLM writing capabilities.

% \clearpage

% \bibliographystyle{acl_natbib}
\bibliographystyle{plain}
\bibliography{acl_latex}

% \section*{References}

% References follow the acknowledgments in the camera-ready paper. Use unnumbered first-level heading for
% the references. Any choice of citation style is acceptable as long as you are
% consistent. It is permissible to reduce the font size to \verb+small+ (9 point)
% when listing the references.
% Note that the Reference section does not count towards the page limit.
% \medskip

% {
% \small

% [1] Alexander, J.A.\ \& Mozer, M.C.\ (1995) Template-based algorithms for
% connectionist rule extraction. In G.\ Tesauro, D.S.\ Touretzky and T.K.\ Leen
% (eds.), {\it Advances in Neural Information Processing Systems 7},
% pp.\ 609--616. Cambridge, MA: MIT Press.

% [2] Bower, J.M.\ \& Beeman, D.\ (1995) {\it The Book of GENESIS: Exploring
%   Realistic Neural Models with the GEneral NEural SImulation System.}  New York:
% TELOS/Springer--Verlag.

% [3] Hasselmo, M.E., Schnell, E.\ \& Barkai, E.\ (1995) Dynamics of learning and
% recall at excitatory recurrent synapses and cholinergic modulation in rat
% hippocampal region CA3. {\it Journal of Neuroscience} {\bf 15}(7):5249-5262.
% }

% \newpage
\section*{NeurIPS Paper Checklist}

\begin{enumerate}

\item {\bf Claims}
    \item[] Question: Do the main claims made in the abstract and introduction accurately reflect the paper's contributions and scope?
    \item[] Answer: \answerYes{} % Replace by \answerYes{}, \answerNo{}, or \answerNA{}.
    \item[] Justification: The methodologies for the contributions mentioned in the abstract and introduction are detailed in the subsections of Section~\ref{sec: writingbench}, with experimental validation provided in Section~\ref{sec: experiment}.

\item {\bf Limitations}
    \item[] Question: Does the paper discuss the limitations of the work performed by the authors?
    \item[] Answer: \answerYes{} % Replace by \answerYes{}, \answerNo{}, or \answerNA{}.
    \item[] Justification: Limitations are discussed in Appendix~\ref{app: limitation}.

\item {\bf Theory assumptions and proofs}
    \item[] Question: For each theoretical result, does the paper provide the full set of assumptions and a complete (and correct) proof?
    \item[] Answer: \answerYes{} % Replace by \answerYes{}, \answerNo{}, or \answerNA{}.
    \item[] Justification: Section~\ref{sec: experiment} and Appendix~\ref{app: experiment_results} provide detailed supporting experiments and analyses for the conclusions drawn in the paper.

    \item {\bf Experimental result reproducibility}
    \item[] Question: Does the paper fully disclose all the information needed to reproduce the main experimental results of the paper to the extent that it affects the main claims and/or conclusions of the paper (regardless of whether the code and data are provided or not)?
    \item[] Answer: \answerYes{} % Replace by \answerYes{}, \answerNo{}, or \answerNA{}.
    \item[] Justification: Section~\ref{ssec: benchmark_construction} of this paper provides a detailed description of the benchmark construction pipline with relevant prompts listed in Appendix~\ref{app: prompts}. The evaluation dataset and evaluation code can be accessed in the GitHub Repo: https://github.com/X-PLUG/WritingBench. The training parameters and datasets for the critic model and writing model are illustrated in Section~\ref{ssec: experiment_settings}, Section~\ref{ssec: writing_model} and Section~\ref{ssec: evaluation_metric}. The models have also been open-sourced on Hugging Face (download links can be found in the GitHub repository). The latest leaderboard evaluation results can be found at Leaderboard: https://huggingface.co/spaces/WritingBench/WritingBench.


\item {\bf Open access to data and code}
    \item[] Question: Does the paper provide open access to the data and code, with sufficient instructions to faithfully reproduce the main experimental results, as described in supplemental material?
    \item[] Answer: \answerYes{} % Replace by \answerYes{}, \answerNo{}, or \answerNA{}.
    \item[] Justification: The benchmark dataset and code have been open-sourced in the GitHub Repo: https://github.com/X-PLUG/WritingBench with detailed instructions provided in README.

\item {\bf Experimental setting/details}
    \item[] Question: Does the paper specify all the training and test details (e.g., data splits, hyperparameters, how they were chosen, type of optimizer, etc.) necessary to understand the results?
    \item[] Answer: \answerYes{} % Replace by \answerYes{}, \answerNo{}, or \answerNA{}.
    \item[] Justification: The benchamrk evaluation settings and training parameters and datasets for the critic model and writing model are provided in Section~\ref{ssec: experiment_settings}, Section~\ref{ssec: writing_model} and Section~\ref{ssec: evaluation_metric}.

\item {\bf Experiment statistical significance}
    \item[] Question: Does the paper report error bars suitably and correctly defined or other appropriate information about the statistical significance of the experiments?
    \item[] Answer: \answerYes{} % Replace by \answerYes{}, \answerNo{}, or \answerNA{}.
    \item[] Justification: We provide standard deviation for benchmark evaluation in Table~\ref{tab:main table}. 

\item {\bf Experiments compute resources}
    \item[] Question: For each experiment, does the paper provide sufficient information on the computer resources (type of compute workers, memory, time of execution) needed to reproduce the experiments?
    \item[] Answer: \answerYes{} % Replace by \answerYes{}, \answerNo{}, or \answerNA{}.
    \item[] Justification: The compute resources are provided in Section~\ref{ssec: experiment_settings}.
    
\item {\bf Code of ethics}
    \item[] Question: Does the research conducted in the paper conform, in every respect, with the NeurIPS Code of Ethics \url{https://neurips.cc/public/EthicsGuidelines}?
    \item[] Answer: \answerYes{} % Replace by \answerYes{}, \answerNo{}, or \answerNA{}.
    \item[] Justification: The research conducted in the paper conforms with the NeurIPS Code of Ethics.

\item {\bf Broader impacts}
    \item[] Question: Does the paper discuss both potential positive societal impacts and negative societal impacts of the work performed?
    \item[] Answer: \answerYes{} % Replace by \answerYes{}, \answerNo{}, or \answerNA{}.
    \item[] Justification: The impacts are discussed in Appendix~\ref{app: impact}.
    
\item {\bf Safeguards}
    \item[] Question: Does the paper describe safeguards that have been put in place for responsible release of data or models that have a high risk for misuse (e.g., pretrained language models, image generators, or scraped datasets)?
    \item[] Answer: \answerYes{} % Replace by \answerYes{}, \answerNo{}, or \answerNA{}.
    \item[] Justification: Our benchmark serves exclusively for evaluation purposes. All released prompts undergo manual review to eliminate harmful information. 

\item {\bf Licenses for existing assets}
    \item[] Question: Are the creators or original owners of assets (e.g., code, data, models), used in the paper, properly credited and are the license and terms of use explicitly mentioned and properly respected?
    \item[] Answer: \answerYes{} % Replace by \answerYes{}, \answerNo{}, or \answerNA{}.
    \item[] Justification: The assets used in the paper have been properly cited.

\item {\bf New assets}
    \item[] Question: Are new assets introduced in the paper well documented and is the documentation provided alongside the assets?
    \item[] Answer: \answerYes{} % Replace by \answerYes{}, \answerNo{}, or \answerNA{}.
    \item[] Justification: We have fully discussed the benchmark in this paper and provided instructions and terms of use in GitHub Repo: https://github.com/X-PLUG/WritingBench.

\item {\bf Crowdsourcing and research with human subjects}
    \item[] Question: For crowdsourcing experiments and research with human subjects, does the paper include the full text of instructions given to participants and screenshots, if applicable, as well as details about compensation (if any)? 
    \item[] Answer: \answerYes{} % Replace by \answerYes{}, \answerNo{}, or \answerNA{}.
    \item[] Justification: The instructions and compensation for human annotators are provided.

\item {\bf Institutional review board (IRB) approvals or equivalent for research with human subjects}
    \item[] Question: Does the paper describe potential risks incurred by study participants, whether such risks were disclosed to the subjects, and whether Institutional Review Board (IRB) approvals (or an equivalent approval/review based on the requirements of your country or institution) were obtained?
    \item[] Answer: \answerYes{} % Replace by \answerYes{}, \answerNo{}, or \answerNA{}.
    \item[] Justification: We have adhered to local laws regarding ethical approval. All annotators involved are employed by a professional annotation company, and they have been informed of the data usage and provided consent.

\item {\bf Declaration of LLM usage}
    \item[] Question: Does the paper describe the usage of LLMs if it is an important, original, or non-standard component of the core methods in this research? Note that if the LLM is used only for writing, editing, or formatting purposes and does not impact the core methodology, scientific rigorousness, or originality of the research, declaration is not required.
    %this research? 
    \item[] Answer: \answerYes{} % Replace by \answerYes{}, \answerNo{}, or \answerNA{}.
    \item[] Justification: The involvement of LLMs in the benchmark construction process and scoring is described in Section~\ref{ssec: benchmark_construction} and Section~\ref{ssec: evaluation_metric}.

\end{enumerate}

%%%%%%%%%%%%%%%%%%%%%%%%%%%%%%%%%%%%%%%%%%%%%%%%%%%%%%%%%%%%

\newpage
\appendix

% \newpage
\section{Experiment Results}
\label{app: experiment_results}

\subsection{Validation of Domain Taxonomy Construction and Requirement Dimension}
\label{app: domain_taxonomy_construction}

With existing benchmarks focusing on limited tasks (e.g. fiction), the domain taxonomy of our benchmark is rigorously designed to reflect real-world writing scenarios. We leverage the industrial background of team members to refine the domain taxonomy using over 200K anonymized real-user writing query data under strict data security protocols. Our initial two-tier domain system is established, referencing the mature workflows of industry product teams, with an additional ``Other'' category under each primary domain. This is iteratively refined through multiple rounds: in each iteration, we use the prompt outlined in Appendix~\ref{app: query_classification} and perform classification using GPT-4o to tag 2K randomly selected writing queries data according to the current subdomain labels, with each query potentially receiving multiple tags. Subsequently, human annotation validate the tags and adjust the domain settings for the next iteration, aiming to reduce overlaps and minimize the ``Other'' category. The final tags achieve stable after three iterations, ensuring broad domain coverage.

To evaluate the classification accuracy of GPT-4o, we validate the effectiveness of the model’s automatic annotations on a writing task tag system provided by the production team. This tag system consists of 8 primary domains and 44 subdomains, with a dataset of 3,000 entries that have been manually annotated and verified under strict data security protocols. The recall rate is calculated such that if a manually assigned tag is found within the list of tags generated by the model, it is considered successfully retrieved. The overall recall rate for primary domains is approximately 96\%, while the recall rate for subdomains is about 93\%.

By analyzing the queries provided by the production team, we identify three prevalent requirement dimensions: style, format, and length. To validate these categories, we perform model-based annotations across several iterations, using the same prompt outlined in Appendix~\ref{app: query_classification}. After multiple rounds of analysis, the average frequencies are determined as follows: style-related requirements account for approximately 29.4\%, format-related requirements comprise about 22.48\%, and length-related requirements make up around 20.8\%. The remaining requirements largely correspond to criteria closely associated with writing materials or other demands that are challenging to generalize into abstract dimensions.

\subsection{Ablation of CoT in Creative content}
\label{app:creative}

To validate the impact of CoT reasoning on creative content generation, we conduct SFT experiments on filtered subset described in Section~\ref{ssec: writing_model} on Qwen-2.5-32B-Instruct. Two variants are developed: 1) A base model trained with CoT-formatted instructions, adhering to DeepSeek-R1's original output format, and 2) an ablated version (-w/o CoT) trained using only the response content. These models are evaluated on the Literature \& Art (D4) subset of WritingBench and EQBench, a specialized benchmark for creative writing evaluation, to thoroughly examine their capabilities in generating creative content.

\vspace{2em}
\begin{table}[ht]
  \centering
  \small
  \caption{Ablation of CoT in creative content on two benchmarks.}
  \begin{tabular}{l|cc}
    \toprule
    \textbf{Models} & \textbf{WritingBench-D4} & \textbf{EQBench}  \\
    \midrule
    Deepseek-R1 & 7.70 & 84.99  \\
    Qwen-2.5-32B-Instruct & 5.59 & 48.17  \\
    % Qwen-Max & 8.39 & 79.53 \\
    \midrule
    Qwen-2.5-32B-CoT & 7.58 & 82.48  \\
    - w/o CoT & 7.54 & 79.43 \\   
    % \textbf{- w/ CoT} & \textbf{8.66} & \underline{82.48}  \\
    \bottomrule
  \end{tabular}
  \label{tab:creative}
\end{table}
\vspace{1em}

As shown in Table~\ref{tab:creative}, the CoT-enhanced model outperform the non-reasoning ones on both WritingBench-D4 and EQBench, showing CoT's effectiveness in generating creative content.

\clearpage

\subsection{Human Consistency}
\label{app:human_consistency}

% \subsubsection{Human Preference Annotation}
% \label{app: human_consistency_annotation}

\begin{figure}[h]
  \centering
  \includegraphics[width=1\textwidth]{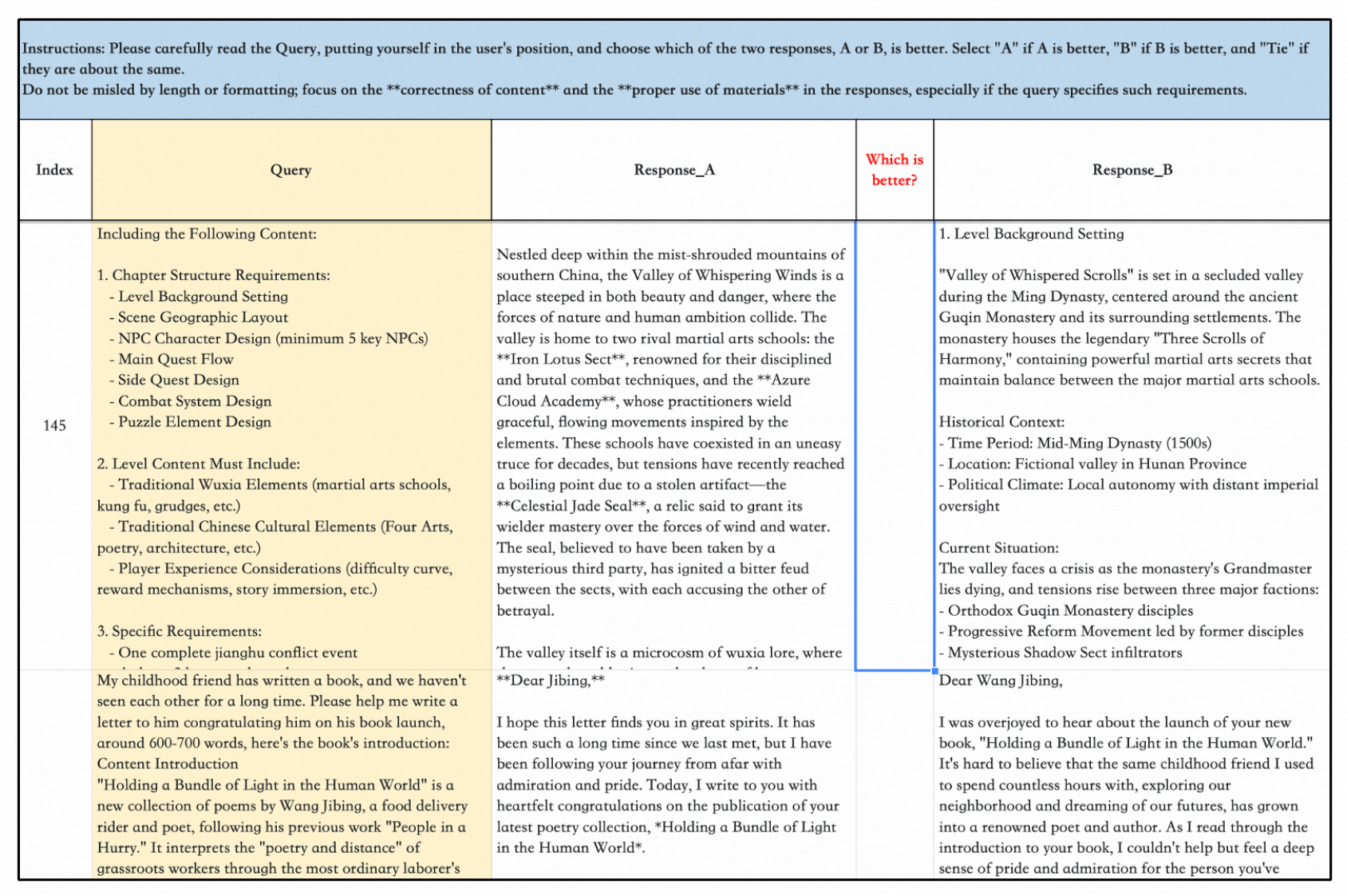}
  \caption{Annotation Interface.}
  \label{fig:interface}
\end{figure}

Human evaluations are conducted via DingTalk Docs (a cloud-based collaborative platform\footnote{\url{https://www.dingtalk.com/en}}), where annotators access standardized spreadsheets. The spreadsheets contain triplets of <Query, Response\_A, Response\_B> with queries highlighted in yellow and instructions are provided in blue header sections (refer to the interface screenshot in Figure~\ref{fig:interface}). To mitigate positional bias, response ordering is randomized per instance, and annotators could zoom into cells for full-text review. The interface enable keyword searches across responses and queries for systematic comparison. The test set excludes inputs exceeding 5,000 tokens to reduce cognitive load, with explicit instructions emphasizing content quality over text length or surface formatting. Five linguists achieved substantial agreement ($\kappa = 0.69$), demonstrating rigorous annotation reliability.

\vspace{2em}
\vspace{3pt}

\subsection{Ablation of Length}
\label{app:lenth_exp}

\begin{figure}[h]
  \centering
  \includegraphics[width=1\linewidth]{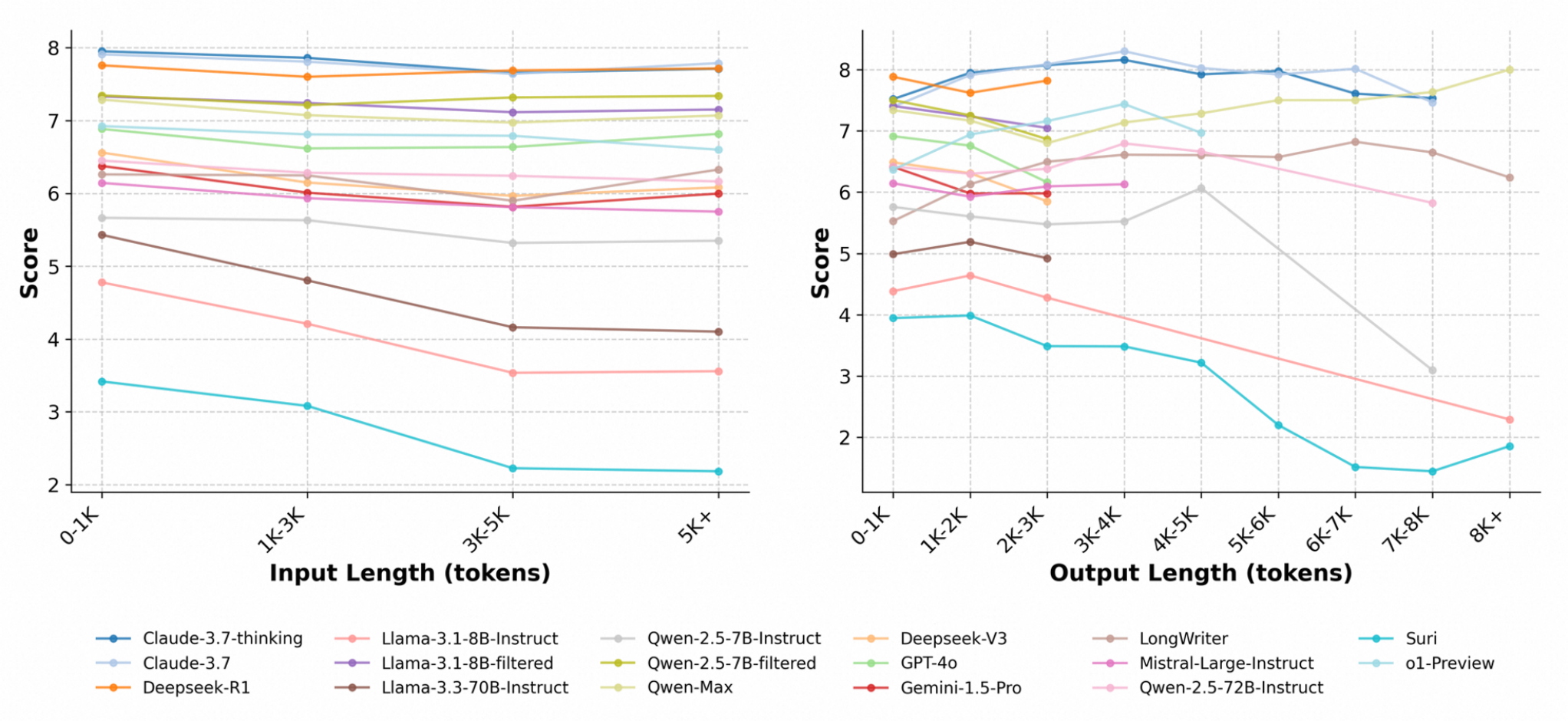}
  \caption{Scores of different models across various input and output lengths on the WritingBench.}
  \label{app:length}
\end{figure}

We assess the performance of LLMs across varying input and output lengths (see Figure~\ref{app:length}), with statistical validity ensured by excluding intervals containing fewer than 5 samples. Experiments on input length reveal that most SOTA models generally maintain consistent performance regardless of input length variations, attributable to their advanced long-context comprehension capabilities.
However, analysis on output length shows that some models exhibit inherent limitations in response generation length, typically producing outputs constrained to approximately 3,000 tokens. Small models, such as Suri, Qwen-2.5-7B-Instruct, Llama-3.1-8B-Instruct, suffer more performance degradation characterized by repetitive outputs. Notably, Claude-3.7 and its reasoning model, Qwen-Max and LongWriter effectively support extended response lengths. These findings highlight the importance of improving long-output generation capabilities and optimizing length handling in writing tasks.

\vspace{20em}
\subsection{Case Study}

\subsubsection{Bad Case of Length Constraint}
Current models often face challenges when handling length-related requirements. Figure~\ref{app:length bad case} illustrates bad cases where the LLM struggle with adhering to length constraints.

\begin{table}[H]
    \renewcommand{\arraystretch}{1.3}
    %\vspace{0.5em}
    \centering
    \caption{\label{app:length bad case} Bad case of length requirement. Query from WritingBench with index 539, response generated by Gemini-1.5-Pro.}
    \vspace{0.5em}
    \begin{adjustbox}{width=0.8\columnwidth}
        \begin{tabularx}{\columnwidth}{X}
            \hline
            \rowcolor{gray!20} % Set the background color to light gray
            \textbf{Query} \\
            \hline
            Create advertising slogans for a pet grooming salon, including: \\
            - Main slogan (10-15 words) \\
            - Sub-slogan (15-20 words) \\
            - Service tagline (8-12 words) \\
            - Brand promise (10-15 words) \\
            The slogans should reflect professional pet grooming and care services while emphasizing thoughtful care for pet owners. The slogans should be catchy, memorable, with warm and friendly wording, and a light, cheerful tone. \\
            \hline
            \rowcolor{gray!20} % Set the background color to light gray
            \textbf{Response} \\
            \hline
            **Main Slogan:** Pamper Your Pet, Delight Your Heart. \\
            **Sub-Slogan:** Expert grooming and loving care, making your furry friend feel fabulous, inside and out. \\
            **Service Tagline:** Gentle paws, happy pets, stress-free smiles. \\
            **Brand Promise:** Quality grooming, compassionate care, every time. \\
            \hline
        \end{tabularx}
    \end{adjustbox}
\end{table}

% \begin{figure}[h]
%   \centering
%   \includegraphics[width=1\linewidth]{pics/input_length.pdf}
% \caption{Scores of different models across various input lengths on the WritingBench.}  \label{app:input_length}
% \end{figure}

% \begin{figure}[h]
%   \centering
%   \includegraphics[width=1\linewidth]{pics/output_length.pdf}
%   \caption{Scores of different models across various output lengths on the WritingBench.}
%   \label{app:output_length}
% \end{figure}

\clearpage
\subsubsection{Criteria Generation Comparison}
\label{app:criteria_comparison}

\begin{figure}[h]
  \centering
  \includegraphics[width=1\textwidth]{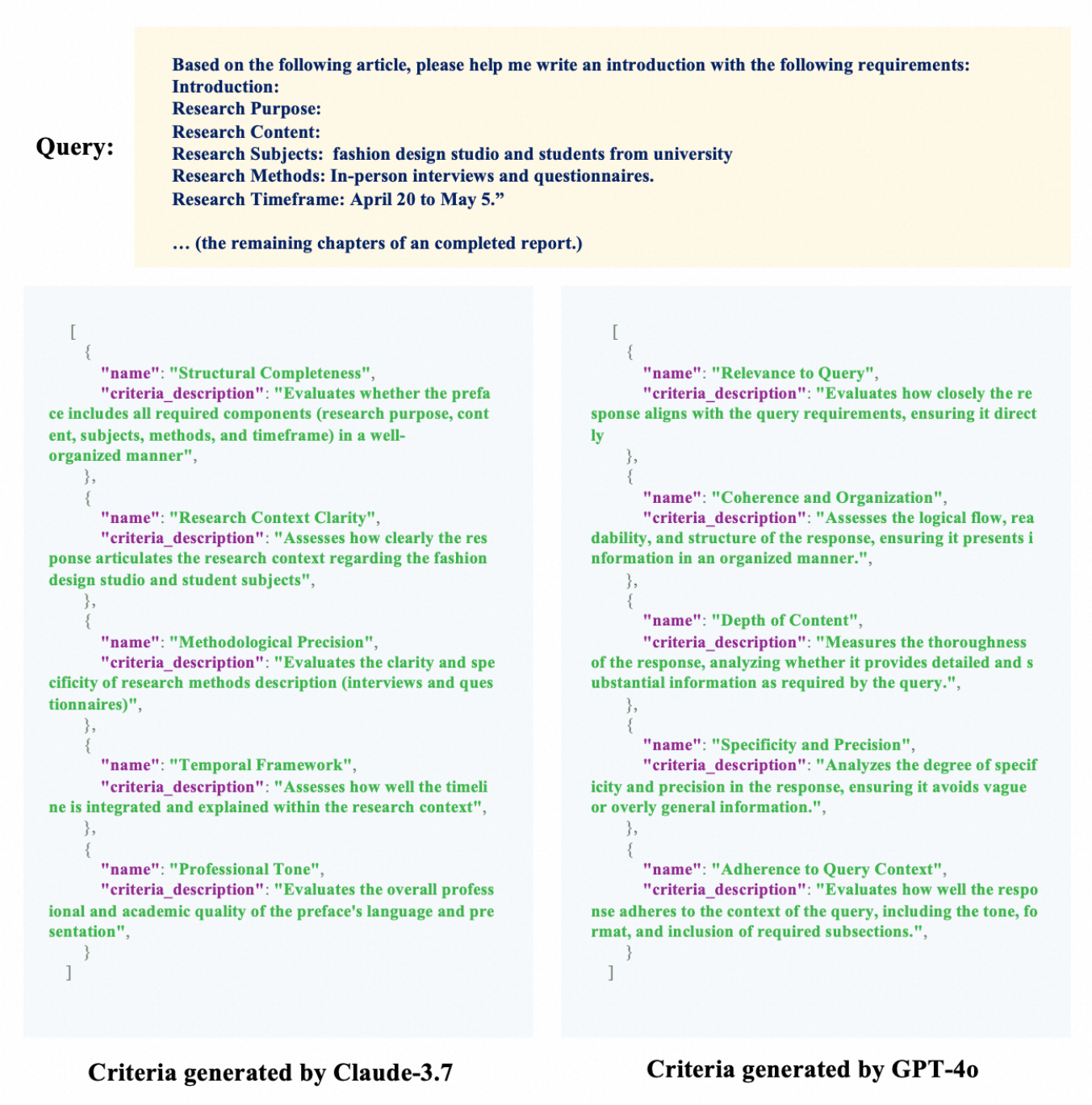}
  \caption{Compasion of criteria generated by Claude-3.7 and GPT-4o.}
  \label{fig:criteria comparison}
\end{figure}

We compare the ability of various models to generate criteria and ultimately selected Claude-3.7, which demonstrates advantages in diversity, comprehensiveness, and rationality of the criteria. Figure~\ref{fig:criteria comparison} presents a comparison between Claude-3.7 and GPT-4o based on the same example query. The criteria descriptions generated by Claude-3.7 show a higher degree of integration with the specific requirements of the query. In contrast, we observe in more cases that GPT-4o tends to uses similar criteria and has a lower level of integration with the query, such as not adequately considering information about the material.

\vspace{2em}
\vspace{3pt}

\clearpage
\section{Benchmark Statistics}
\label{app: Benchmark Statistics}

\subsection{Overview of Six Primary Domains}
\label{app: domains}

\begin{enumerate}

\item \textbf{Academic \& Engineering:} This domain encompasses the structured and formalized nature of academic writing workflows, focusing on clarity, precision, and adherence to rigorous standards. includes the creation of paper outlines, abstracts, literature reviews, experiment reports, and technical documents such as patents and test reports. The writing prioritizes logical argumentation, thorough analysis, and the integration of empirical evidence.

\item \textbf{Finance \& Business:} Writing in this domain is analytical and strategic, aimed at informing decision-making and promoting corporate objectives. It includes a wide range of documentation such as contracts, market analyses, investment reports, strategic plans, and operational materials like product specifications and sales reports. The emphasis is on clarity and conciseness, with a focus on financial acumen and strategic insights.

\item \textbf{Politics \& Law:} This domain demands an authoritative and formal tone, as it involves the composition of government documents, legal writings, and political communications. These materials require a careful balance between clarity and formality, often employing complex and structured language. The aim is to clearly convey policy positions, legal arguments, or political messages while strictly adhering to legal and procedural standards. 

\item \textbf{Literature \& Art:}  This domain covers the creative and expressive realms of writing, including novels, poetry, scripts, artistic designs, and critiques of books and movies. Writers explore thematic and emotional depths, crafting works that connect with audiences on a human level. The language is rich and evocative, allowing for a personal exploration of ideas that engage and move the reader.

\item \textbf{Education:} This domain involves pedagogical materials and educational communication, including lesson plans, course designs, feedback, assignments, and institutional communications like admissions promotions and parent-teacher meeting scripts. The writing prioritizes clarity, accessibility, and instructional effectiveness, using an approachable tone to facilitate learning and engagement.

\item \textbf{Advertising \& Marketing:} Writing in this domain is vibrant and persuasive, designed to captivate and influence target audiences across various digital platforms. It includes social media scripts, advertising copy, brand narratives, and multimedia campaign materials. The writing is dynamic and strategic, with a creative twist, necessitating a deep understanding of audience psychology and trend dynamics to ensure that the content is appealing and strategically effective.
\end{enumerate}

\subsection{Overview of 100 Secondary Subdomains}
See Table~\ref{tab:writing_domains1}, Table~\ref{tab:writing_domains2} and Table~\ref{tab:writing_domains3}.

\clearpage

\begin{table*}
\centering
\footnotesize
\setlength{\extrarowheight}{0pt}
\addtolength{\extrarowheight}{\aboverulesep}
\addtolength{\extrarowheight}{\belowrulesep}
\setlength{\aboverulesep}{0pt}
\setlength{\belowrulesep}{0pt}
% \refstepcounter{table}
\caption{Subdomains in Academic \& Engineering and Finance \& Business.}

\begin{tabular}{l|l}
\toprule
\textbf{SubDomain} & \multicolumn{1}{c}{\textbf{Description}} \\
\midrule

\multicolumn{2}{c}{{\cellcolor[rgb]{0.85,0.85,0.85}}\textit{Academic \& Engineering}} \\
\midrule
Paper Outline & Hierarchical organization of research components and logical flow \\
Acknowledgments & Formal recognition of institutional and individual support \\
Limitations & Systematic identification of methodological constraints and scope boundaries \\
Defense Presentation & Presentation supporting materials, such as slides \\
Research Proposal & Investigation blueprint with validation road map \\
Technical Documentation & Implementation specifications and system interface protocols \\
Experiments & Parameterized validation framework with controlled variable analysis \\
Introduction & Contextual foundation establishing research gaps and significance \\
Conclusion & synthesize the main findings of the research or project \\
Test Report & Evaluations of testing activities and performance \\
Contributions & Novel aspects differentiating the work from prior research \\
Internship Report & Chronological documentation of a practical work placement \\
Literature Review & Critical gap analysis through scholarly works taxonomy \\
Defense Script & Oral presentations and responses for research defense. \\
Abstract & Summary of research objectives, methods, results, and significance \\
Engineering Report & Technical analysis on tasks, methodologies, and outcomes \\
Patent & Legal-technical specification of novel implementable claims \\
\midrule

\multicolumn{2}{c}{{\cellcolor[rgb]{0.85,0.85,0.85}}\textit{Finance \& Business}} \\
\midrule
Meeting Minutes & Concise documentation of key discussion points, decisions, and action items \\

User Research & Insight collection on user needs and behaviors to inform product or service design \\

Business Correspondence & Formal communication with internal or external stakeholders for business purposes \\

Human Resource Management & Strategies and processes for managing workforce effectively \\

Recruitment & Strategies for attracting, selecting, and onboarding suitable candidates \\

Briefing & Summarized information provided to stakeholders ahead of a task or meeting \\

Event Planning & Coordinated organization of logistics and activities for event execution \\

Market Research & Systematic collection and analysis about market and consumer \\

Market Analysis & Evaluation of market trends, size, competitors, and dynamics \\

Risk Management & Identification, assessment, and prioritization of risks with mitigation strategies \\

Sales Report & Summary of sales activities, performance, and revenue over a specific period \\

Pitch Deck & Visual presentation for communicating business ideas or proposals to investors \\

Contract & Legally binding agreement detailing terms and conditions for business transactions \\

Tender Document & Formal proposal request containing project specifications and bidding instructions \\

Investment Analysis & Evaluation of financial investments to determine potential returns and risks \\

Product Proposal & Detailed plan outlining the development, features, and potential of new products \\

Strategic Planning & Business goal setting with actionable strategies for desired outcomes \\

Financial Reports & Comprehensive statements reflecting the financial performance and status \\

Requirements Specification & Documentation detailing functional and non-functional requirements for a project \\

Bid Proposal & Formal offer to supply goods or services at a set price, meeting client needs \\
\midrule

\bottomrule
\end{tabular}
\label{tab:writing_domains1}
\end{table*}

\begin{table*}
\centering
\footnotesize
\setlength{\extrarowheight}{0pt}
\addtolength{\extrarowheight}{\aboverulesep}
\addtolength{\extrarowheight}{\belowrulesep}
\setlength{\aboverulesep}{0pt}
\setlength{\belowrulesep}{0pt}
% \refstepcounter{table}
\caption{Subdomains in Politics \& Law and Literature \& Art.}

\begin{tabular}{l|l}
\toprule
\textbf{Subdomain} & \multicolumn{1}{c}{\textbf{Description}} \\
\midrule

\multicolumn{2}{c}{{\cellcolor[rgb]{0.85,0.85,0.85}}\textit{Politics \& Law}} \\
\midrule

Legal Opinion & Authoritative assessment and guidance on legal matters or questions \\

Government Speech & Formal address by government officials outlining policies or positions \\

Judgment Document & Official written decision or order issued by a court \\

Legal Agreement & Binding contract setting out terms and obligations between parties \\

Case Study & In-depth analysis of a legal case for educational or professional purposes \\

Case Bulletin & Summary and update on ongoing or concluded legal cases \\

Legal Consultation & Professional advice provided on legal rights, responsibilities, or strategies \\

Regulatory Analysis & Examination of rules and regulations affecting compliance and enforcement \\

Meeting Summary & Brief overview of discussions, decisions, and outcomes from a meeting \\

Ideological Report & Analysis or commentary on political or ideological trends and perspectives \\

Policy Interpretation & Explanation or clarification for public or organizational guidance \\

Official Document & Formal written record issued by government entities or officials \\

Legal Awareness Campaign & Initiative to educate the public on legal rights and responsibilities \\

Defense Plea & Formal written argument submitted by the defense in a legal proceeding \\

Party Membership Application & Form and process for joining a political party \\

Policy Advocacy & Efforts to influence or promote specific policy changes or implementations \\

Work Report & Report of activities, achievements, and challenges within a specific period \\

Deed Achievement & Record highlighting significant accomplishments and contributions \\

Litigation Documents & Legal filings and paperwork submitted in the course of a lawsuit \\

White Paper & Authoritative report providing information or proposals on an issue \\

\midrule
\multicolumn{2}{c}{{\cellcolor[rgb]{0.85,0.85,0.85}}\textit{Literature \& Art}} \\
\midrule

Character Design & Creation and development of detailed characters for stories or visual media \\

Greeting Message & Friendly or formal introductory statement used for various occasions \\

Host Script & Guided narration and dialogue for a presenter during an event or show \\

Novel Outline & Structured plan for the plot, characters, and settings of a novel \\

Podcast Script & Written content outlining the dialogue and segments for podcast episodes \\

Derivative Work & Creative work based on or inspired by an existing piece \\

Reading Reflection & Personal thoughts and analysis on a piece of literature \\

Video Script & Script detailing dialogue and action for video content creation \\

Book Review & Critical evaluation and summary of a book's content and impact \\

Game Design & Creation of mechanics, stories, and interfaces for games \\

Lyric Writing & Crafting of words for songs with rhyme and meter considerations \\

Brainstorm & Rough ideas and notes generated during a creative thinking session \\

Plot Development & Process of mapping out the storyline and narrative structure \\

Prose & Written or spoken language in its ordinary form, without metrical structure \\

Screenplay & Scripted blueprint for film or television with dialogue and directions \\

Novel Manuscript & Complete text of a novel prepared for publication \\

Biography & Detailed account of a person's life experiences and achievements \\

Film/TV Review & Analytical critique of a film or television show's content and effectiveness \\

Poetry & Artistic composition using rhythmic and metaphorical language \\

Fan Fiction & Amateur stories by enthusiasts featuring characters from existing media \\

\midrule

\bottomrule
\end{tabular}
\label{tab:writing_domains2}
\end{table*}

\begin{table*}[]
\centering
\footnotesize
\setlength{\extrarowheight}{0pt}
\addtolength{\extrarowheight}{\aboverulesep}
\addtolength{\extrarowheight}{\belowrulesep}
\setlength{\aboverulesep}{0pt}
\setlength{\belowrulesep}{0pt}
% \refstepcounter{table}
\caption{Subdomains in Education and Advertising \& Marketing.}

\begin{tabular}{l|l}
\toprule
\textbf{SubDomain} & \multicolumn{1}{c}{\textbf{Description}} \\
\midrule

\multicolumn{2}{c}{{\cellcolor[rgb]{0.85,0.85,0.85}}\textit{Education}} \\
\midrule

Training Reflection & Personal assessment of training experiences and learned insights \\

Class Activity & Planned exercises or tasks designed to engage students in learning \\

Parent-Teacher Meeting & Formal discussion between educators and parents about student progress \\

Lesson Plan & Structured outline of educational objectives and teaching methods \\

Teaching Materials & Resources used to aid in presenting information to students \\

Assignment Grading & Evaluation and scoring of student work based on specific criteria \\

Curriculum Design & Development of educational content, structure, and delivery methods \\

Educational Report & Analysis or summary of educational outcomes and performance \\

Coursework & Academic work assigned to students as part of a course \\

Evaluation Comments & Feedback provided on student performance and areas of improvement \\

Educational Consulting & Professional guidance on educational strategies and systems \\

Admissions Promotion & Strategies and activities to encourage enrollment in educational institutions \\

\midrule

\multicolumn{2}{c}{{\cellcolor[rgb]{0.85,0.85,0.85}}\textit{Advertising \& Marketing}} \\
\midrule

Sales Letter & Persuasive written communication intended to motivate potential buyers \\

Product Description & Detailed overview of a product’s features, benefits, and uses \\

Social Media Content & Engaging text, images, or videos crafted for online platforms \\

Multimedia Script & Planned screenplay integrating various forms of media for marketing \\

Promotional Copy & Compelling text written to boost interest and sales of products \\

Promotional Voiceover & Recorded narration to accompany marketing visuals or ads \\

Travel Guide & Informative content offering insights and tips for travelers \\

Brand Story & Narrative that outlines the history, values, and mission of a brand \\

Personal Blog & Individual commentary or stories shared in an informal online format \\

Marketing Commentary & Analytical thoughts on marketing trends and strategies \\

Slogans & Catchy and memorable phrases designed to convey brand identity \\

\bottomrule
\end{tabular}
\label{tab:writing_domains3}
\end{table*}

\clearpage 

\section{Prompts}
\label{app: prompts}

\subsection{Query Classification Prompt}
\label{app: query_classification}

Introduced in Appendix~\ref{app: domain_taxonomy_construction}.

\begin{tcolorbox}[colback=blue!5,
    colframe=blue!80,
    width=14cm, 
    arc=2mm, auto outer arc, 
    title={\textbf{Classification System Prompt}}, 
    breakable, enhanced jigsaw, 
    before upper={\parindent15pt\noindent}]
You are an expert in query analysis.
\end{tcolorbox}

% \vspace{0.5cm} % Space between boxes
During each iteration, Put current secondary domain tags under ** Domains **. For example, "1. Academic \& Engineering" represents a primary domain, while "1a. Thesis Outline" represents a subdomain.

\begin{tcolorbox}[colback=blue!5, 
    colframe=blue!80,
    width=14cm, 
    arc=2mm, auto outer arc, 
    title={\textbf{Query Classification Prompt}}, 
    breakable, enhanced jigsaw, 
    before upper={\parindent15pt\noindent}]
Please determine which of the following domains the query belongs to and identify any stylistic, formatting, or length requirements. \\
\newline
**Example of Format Requirement**: Mimicking the format of uploaded documents, adhering to a given outline format, conforming to academic paper formatting standards, etc.  \\
**Example of Style Requirement**: Suitable for children’s reading, rigorous language, humorous tone, etc.  \\
**Example of Length Requirement**: Word count, duration, or other constraints related to output size. \\
\newline
**  Query ** \\
\{query\} \\
\newline
** Domains **\\
1. Academic \& Engineering \\
\hspace*{1em}1a. Thesis Outline \\
\hspace*{1em}1b. Literature Review \\
\hspace*{1em}...\\
\hspace*{1em}1r. Others related to Academic \& Engineering \\
\newline
2. Finance \& Business \\
\hspace*{1em}2a. Market Research \\
\hspace*{1em}2b. Sales Report \\
\hspace*{1em}... \\
\hspace*{1em}2l. Others related to Finance \& Business \\
... \\
7. Other \\
\hspace*{1em}7a. Others \\
\newline
** Output format ** \\
Return in JSON, strictly output according to the following format, do not output other content \\
\{ \\
\hspace*{1em}"domain": ["xx.yyy",...] // Domains involved, such as "6c. Marketing Letter", "8d. Educational Consulting", if the data is invalid and cannot be determined, return an empty list. \\
\hspace*{1em}""style": "",  // style requirement if present (e.g. "academic format"), else empty string \\
\hspace*{1em}""format": "",  // format specification if present (e.g. "child-friendly tone"), else empty string \\
\hspace*{1em}""length": ""  // length constraint (e.g. "500 words") or empty string \\
\} \\
\end{tcolorbox}

\subsection{Initial Query Generation Prompt}
\label{app:InitialQueryGenerationPrompt}
Introduced in Section~\ref{sssec: model-augmented_query_generation}. You can specify the language of the generated query in the prompt.
\begin{tcolorbox}[colback=blue!5,
    colframe=blue!80, 
    width=14cm, 
    arc=2mm, auto outer arc, 
    title={\textbf{Query Classification Prompt}},
    breakable, enhanced jigsaw, 
    before upper={\parindent15pt\noindent}]
    Generate \{NUM\} different writing requests under \texttt{\{subdomain\}} within the context of \texttt{\{primary\_domain\}} in English / Chinese. Ensure the requests are as detailed and specific as possible, and reflect realistic user tone and needs. \\
    \newline
    Please return in the following JSON format, and do not include anything outside of JSON: \newline
    [ \newline
    \hspace*{1em}"Writing request 1",  \newline
    \hspace*{1em}"Writing request 2",  \newline
    \hspace*{1em} \dots \newline
    ] \\ 
\end{tcolorbox}

\subsection{Guidance Pool}
\label{app:GuidancePool}

Introduced in Section~\ref{sssec: model-augmented_query_generation}. Randomly select 0-6 items each time.
\begin{tcolorbox}[colback=blue!5,
    colframe=blue!80, 
    width=14cm, 
    arc=2mm, auto outer arc, 
    title={\textbf{Query Refinement Guidance Pool}},
    breakable, enhanced jigsaw, 
    before upper={\parindent15pt\noindent}]
    [ \newline
    \hspace*{1em} "Add a requirement for generating specific lengths.",\newline
    \hspace*{1em} "Include format adherence requirements, such as writing according to a prescribed outline or outputting in a specific format.",\newline
    \hspace*{1em} "Add style requirements, like drafting a speech suitable for a particular occasion or adopting the style suitable for a specific audience or mimicking a particular tone.",\newline
    \hspace*{1em} "Incorporate user personalization needs, such as considering the user's identity or integrating personal experiences.",\newline
    \hspace*{1em} "Include more specific content requirements, like details about a particular event or focusing on specific content.",\newline
    \hspace*{1em} "Express concisely in one sentence."\newline
    ] \\            
\end{tcolorbox}

\subsection{Query Refine Prompt}
\label{app:QueryRefinePrompt}

Introduced in Section~\ref{sssec: model-augmented_query_generation}.
\begin{tcolorbox}[colback=blue!5,
    colframe=blue!80, 
    width=14cm, 
    arc=2mm, auto outer arc, 
    title={\textbf{Query Refinement Prompt}},
    breakable, enhanced jigsaw, 
    before upper={\parindent15pt\noindent}]
    Please refine and enhance the original writing requirements in the context of generating content in \texttt{\{domain2\}} from \texttt{\{domain1\}} based on the provided guidance. Include as many details as possible and indicate whether additional writing materials are needed. \newline
    \newline
    ** Original Writing Requirements **\newline
    \texttt{\{query\}} \newline
    \newline
    ** Guidance for Modification **\newline
    \texttt{\{guidance\}} \newline
    \newline
    ** Output Requirements **\newline
    Return the result strictly in the following JSON format, with no additional content outside the JSON: \newline
    \{ \\
    \hspace*{1em} "query": "Modified writing requirements", \newline
    \hspace*{1em} "material": "Whether additional reference materials are needed to supplement the writing requirements. If needed, provide suggestions for the materials; if not needed, return" \newline
    \} \\
\end{tcolorbox}

\subsection{Criteria Generation Prompt}
\label{app:criteria generation prompt}

Introduced in Section~\ref{ssec: evaluation_metric}.

\begin{tcolorbox}[colback=blue!5,
    colframe=blue!80, 
    width=14cm, 
    arc=2mm, auto outer arc, 
    title={\textbf{Evaluation System Prompt}},
    breakable, enhanced jigsaw, 
    before upper={\parindent15pt\noindent}]
    You are an expert evaluator with extensive experience in evaluating the response of a given query. \\
\end{tcolorbox}

\vspace{0.5cm}

\begin{tcolorbox}[colback=blue!5,
    colframe=blue!80, 
    width=14cm, 
    arc=2mm, auto outer arc, 
    title={\textbf{Criteria Generation Prompt}},
    breakable, enhanced jigsaw, 
    before upper={\parindent15pt\noindent}]
    Please generate five strict evaluation criteria for assessing the response given the following query. Each criterion should include the following fields: name, criteria\_description, 1-2, 3-4, 5-6, 7-8, 9-10.
    \newline
    The criteria should be designed to emphasize detailed assessment and distinguish subtle differences in quality. Ensure that the criteria can discern issues such as relevance, coherence, depth, specificity, and adherence to the query context. 
    \newline
    Do not include any additional text. Only output the criteria in the specified JSON format. \newline
    \newline
    ** Query ** \newline
    \texttt{\{query\}} \newline
    \newline
    ** Output format ** \newline
    [ \newline
    \hspace*{1em}\{ \newline
    \hspace*{2em}"name": "first\_criteria\_name", \newline
    \hspace*{2em} "criteria\_description": "Description for the first criteria, emphasizing detailed and critical assessment.", \newline
    \hspace*{2em} "1-2": "Low score description: Critical deficiencies and major issues that prevent adequate functionality.", \newline
    \hspace*{2em} "3-4": "Below average score description: Lacking with noticeable shortcomings that impact overall effectiveness and require improvement.", \newline
    \hspace*{2em} "5-6": "Average score description: Adequate but not exemplary, Baseline performance that meets essential requirements. Most models may achieve this score.", \newline
    \hspace*{2em} "7-8": "Above average score description: Strong performance characterized by competent execution, though minor refinements are needed to achieve excellence.", \newline
    \hspace*{2em} "9-10": "High score description: Exceptional performance with all aspects optimally addressed, demonstrating superior effectiveness and quality without any flaws." \newline
    \hspace*{1em}\}, \newline
    \hspace*{1em}... \newline
    ] \\%\newline
\end{tcolorbox}

\clearpage
\subsection{Rubric-based Scoring Prompt}
\label{app:rubric-based scoring prompt}

Introduced in Section~\ref{ssec: evaluation_metric}.
\begin{tcolorbox}[colback=blue!5,
    colframe=blue!80, 
    width=14cm, 
    arc=2mm, auto outer arc, 
    title={\textbf{Evaluation System Prompt}},
    breakable, enhanced jigsaw, 
    before upper={\parindent15pt\noindent}]
    You are an expert evaluator with extensive experience in evaluating the response of a given query. \\
\end{tcolorbox}

Since the query and response may be very long, the \{criteria\} will appear twice: once before and once after the query and response.

\begin{tcolorbox}[colback=blue!5,
    colframe=blue!80, 
    width=14cm, 
    arc=2mm, auto outer arc, 
    title={\textbf{Scoring Prompt}},
    breakable, enhanced jigsaw, 
    before upper={\parindent15pt\noindent}]
    Evaluate the Response based on the Query and Criteria provided following the Scoring Rules. \newline
    \newline
    ** Scoring Rules ** \newline
    "1-2": "Low score description: Critical deficiencies and major issues that prevent adequate functionality." \newline
    "3-4": "Below average score description: Lacking with noticeable shortcomings that impact overall effectiveness and require improvement." \newline
    "5-6": "Average score description: Adequate but not exemplary, Baseline performance that meets essential requirements. Most models may achieve this score." \newline
    "7-8": "Above average score description: Strong performance characterized by competent execution, though minor refinements are needed to achieve excellence." \newline
    "9-10": "High score description: Exceptional performance with all aspects optimally addressed, demonstrating superior effectiveness and quality without any flaws." \newline
    \newline
    -Provide reasons for each score by indicating specific strengths or deficiencies within the Response. Reference exact text passages to justify the score, ensuring that each reason is concrete and aligns with the criteria requirements while highlighting key gaps from the ideal answer. \newline
    \newline
    -Be very STRICT and do not be misled by format or length; ensure that the Response is thoroughly evaluated beyond superficial appearances. \newline
    \newline
    -Carefully discern whether the content of the Response is an illusion, appearing substantial but actually entirely fabricated. \newline
    \newline
    -Sometimes the model may only provide an introduction or an overview without truly completing the query, which should be considered a failed response. Carefully discern this. \newline
    \newline
    -Scoring Range: Assign an integer score between 1 to 10 \newline
    \newline
    ** Output format ** \newline
    (Remove symbols that interfere with JSON parsing, don't use " inside reason) \newline
    Return the results in the following JSON format, Only output the following JSON format and nothing else: \newline
    \{ \newline
    \hspace*{1em} "score": an integer score between 1 to 10, \newline
    \hspace*{1em} "reason": "Specific and detailed justification for the score using text elements." \newline
    \} \\
    \newline
    ** Criteria ** \newline
    \texttt{\{criteria\}} \newline
    \newline
    ** Query ** \newline
    \texttt{\{query\}} \newline
    \newline
    ** Response ** \newline
    \texttt{\{response\}} \newline
    \newline
    Provide your evaluation based on the criteria restated below: \\
    \texttt{\{criteria\}} \newline
    \newline
    ** Output format ** \newline
    (Remove symbols that interfere with JSON parsing, don't use " inside reason) \newline
    Return the results in the following JSON format, Only output the following JSON format and nothing else: \newline
    \{ \newline
    \hspace*{1em} "score": an integer score between 1 to 10, \\
    \hspace*{1em} "reason": "Specific and detailed justification for the score using text elements." \newline
    \} \\
\end{tcolorbox}

\section{Limitations}
\label{app: limitation}

This work faces three primary limitations that warrant consideration.
First, both the writing model and critic model are primarily trained using conventional SFT approaches, omitting systematic exploration of enhanced optimization strategies. While we demonstrate the partial efficacy of the CoT mechanisms in creative domains, their potential remains unexplored compared to established successes in mathematical reasoning tasks.

Second, our evaluation framework exhibits diminished precision in handling complex multi-dimensional length requirements, including temporal sequencing constraints and section-specific word counts. This limitation underscores the necessity for enhanced scoring methodologies that integrate learned metrics with structured rule-based evaluations to better regulate output specifications.

Third, inherent challenges persist in obtaining reliable pairwise preference annotations for compositional tasks. Despite rigorous annotation protocols, human evaluators inevitably introduce subjective biases when assessing two fair-well responses, particularly regarding narrative preferences and contextual interpretations. While our consensus-building procedures mitigate some variability, absolute alignment with diverse user preferences remains theoretically unattainable.

\section{Impact}
\label{app: impact}

The introduction of WritingBench and its associated evaluation framework presents significant impacts across multiple dimensions of LLM research and application development:

\begin{itemize}[left=0pt]
\item \textbf{Comprehensive AI Writing Evaluation Ecosystem} \
WritingBench’s extensive query set spans a wide range of domains and requirements, making it an ideal resource for evaluating both general-purpose writing skills and domain-specific expertise. Researchers can use the benchmark to assess a model's overall versatility or focus on specific domains, such as storytelling, scientific writing, or business communication. This dual capability allows for nuanced evaluations that cater to both academic research and practical applications, bridging the gap between theoretical advancements and real-world needs. By publicly releasing WritingBench, including its evaluation protocols, criteria generation tools, and critic models, we contribute to greater transparency and reproducibility in the field of LLM research. Researchers can replicate experiments, validate findings, and build upon our work to further refine evaluation methodologies. 
Please note that although every released query is manually reviewed to remove harmful content, the responses generated via API calls depend on the underlying model’s behavior, cannot be guaranteed to be fully safe or accurate, and should not be used for commercial purposes.

\item \textbf{Facilitating Domain-Specific Research and Development} \
With its inclusion of 100 subdomains, WritingBench provides a unique opportunity for domain-specific optimization. For example, industries such as legal writing, medical documentation, or marketing can evaluate model performance on relevant subsets to select the most suitable models for their needs. Furthermore, WritingBench’s query construction strategy can be leveraged to generate diverse queries when specialized evaluation datasets are lacking in a particular field. This capability supports targeted research the development of highly specialized writing tools. We welcome researchers and experts from all fields to join the discussion and collaborate on the development of this benchmark.

\item \textbf{Advancing Reinforcement Learning and Adaptive Evaluation Frameworks} \
The query-dependent evaluation framework introduced in WritingBench opens new avenues for integrating adaptive evaluation methods into reinforcement learning pipelines. The ability to generate instance-specific criteria dynamically allows for precise scoring mechanisms that can be used to train models through reward-based optimization. Additionally, the framework’s capacity to produce pairwise preference data by comparing predicts against adaptive criteria for improving model alignment with human preferences. This approach enhances the quality of generated text and accelerates the development of models capable of handling complex, multi-dimensional writing tasks.
\end{itemize}

%%%%%%%%%%%%%%%%%%%%%%%%%%%%%%%%%%%%%%%%%%%%%%%%%%%%%%%%%%%%

\end{document}